\documentclass[pdflatex,sn-mathphys-num,iicol]{sn-jnl}

\usepackage{graphicx}%
\usepackage{multirow}%
\usepackage{amsmath,amssymb,amsfonts}%
\usepackage{amsthm}%
\usepackage{mathrsfs}%
\usepackage[title]{appendix}%
\usepackage[dvipsnames]{xcolor}
\usepackage{textcomp}%
\usepackage{manyfoot}%
\usepackage{booktabs}%
\usepackage{algorithm}%
\usepackage{algorithmicx}%
\usepackage{algpseudocode}%
\usepackage{listings}%
\usepackage{xspace}
\usepackage{comment}
\usepackage{multirow}
\usepackage{bm}
\usepackage{float}
\usepackage{wrapfig}
\usepackage{pgfplots}
\usepackage{tikz}
\usepackage{cleveref}
\usepackage[numbers]{natbib}

\pgfplotsset{compat=1.18}
\usepgfplotslibrary{fillbetween}
\usetikzlibrary{positioning}
\usetikzlibrary{pgfplots.groupplots}
\usetikzlibrary{calc}
\usetikzlibrary{shapes}
\usetikzlibrary{arrows.meta}

\newcommand{\eg}{\textit{e.g.}}
\newcommand{\ie}{\textit{i.e.}}

\theoremstyle{thmstyleone}%
\theoremstyle{thmstyletwo}%

\theoremstyle{thmstylethree}%

\begin{document}

\title[Model-based Metric 3D Shape and Motion Reconstruction of Wild Bottlenose Dolphins in Drone-Shot Videos]{Model-based Metric 3D Shape and Motion Reconstruction of Wild Bottlenose Dolphins in Drone-Shot Videos}


\author*[1]{\fnm{Daniele} \sur{Baieri}}\email{daniele.baieri@unimib.it}

\author[2]{\fnm{Riccardo} \sur{Cicciarella}}\email{riccardo.cicciarella@iea.uzh.ch}

\author[2]{\fnm{Michael} \sur{Kr\"utzen}}\email{michael.kruetzen@iea.uzh.ch}

\author[3]{\fnm{Emanuele} \sur{Rodol\`a}}\email{rodola@di.uniroma1.it}

\author[4]{\fnm{Silvia} \sur{Zuffi}}\email{silvia.zuffi@cnr.it}

\affil*[1]{\orgdiv{Department of Informatics, Systems and Communication}, \orgname{University of Milano-Bicocca}, \orgaddress{\street{Viale Sarca}, \city{Milan}, \postcode{20126}, \country{Italy}}}

\affil[2]{\orgdiv{Department of Evolutionary Anthropology}, \orgname{University of Zurich}, \orgaddress{\street{Winterthurerstrasse 190}, \city{Zurich}, \postcode{8057}, \country{Switzerland}}}

\affil[3]{\orgdiv{Department of Computer Science}, \orgname{Sapienza University of Rome}, \orgaddress{\street{Viale Regina Elena}, \city{Rome}, \postcode{00161}, \country{Italy}}}

\affil[4]{\orgdiv{IMATI}, \orgname{CNR}, \orgaddress{\street{via A. Corti 12}, \city{Milan}, \postcode{20133}, \country{Italy}}}


\abstract{We address the problem of estimating the metric 3D shape and motion of wild dolphins from monocular video, with the aim of assessing their body condition. While considerable progress has been made in reconstructing 3D models of terrestrial quadrupeds, aquatic animals remain unexplored due to the difficulty of observing them in their natural underwater environment. To address this, we propose a model-based approach that incorporates a transmission model to account for water-induced occlusion. We apply our method to video captured under different sea conditions. We estimate mass and volume, and compare our results to a manual 2D measurements-based method. Additionally, we apply our method to video of captive animals with known ground truth mass. While in our experiments the manual approach is often more accurate, our method demonstrates a distinct advantage when applied to larger specimen. These findings highlight the potential of our method as a scalable and automated alternative for mass and volume estimation of dolphins from monocular video.}




\maketitle

\section{Introduction}\label{sec:introduction}

Body condition and mass are key traits that influence many physiological processes \cite{lauder1981form}. Body condition represents the amount of energy reserves an animal has in relation to its structural size \cite{hanks1981characterization, millar1990fasting} and provides insight into nutritional status and overall physiological health. Body mass influences an animal's metabolic rate, which subsequently impacts its food requirements, growth rate, reproduction, and mortality \cite{kleiber1947body,anderson2009allometry}. Together, body condition and mass are essential for studying ecology, physiology, and how animals respond to environmental stressors \cite{lauder1981form,garland1983scaling,dobson1992body,stevenson2006condition}.

Body condition reflects the energetic demands of maintenance, growth, and reproduction \cite{kleiber1947body,anderson2009allometry}. In females, higher condition is often linked to greater reproductive success, influencing offspring size, growth, and survival \cite{tauson1993effect,taillon2012mother,christiansen2018maternal}. It also reveals the energetic costs of gestation and lactation. For example, lactating southern right whales lose about 25\% of body volume during a breeding season, while humpback and western gray whales show similar declines in condition due to the high demands of nursing \cite{christiansen2016noninvasive,christiansen2018maternal,bradford2012leaner}. 

Body mass is a fundamental characteristic shaping animal physiology, ecology, and behavior \cite{schmidt1984scaling,calder1982pace}. It directly affects metabolic rates, energy requirements, growth, thermoregulation, aerobic dive limits, and home range size \cite{kleiber1947body,porter2009size,kooyman1981physiology,mcnab1963bioenergetics}. Allometric equations relating body mass to metabolic rate capture much of the inter-specific variation in energy use and allow estimation of bioenergetic needs even without direct measurements \cite{white2011allometric}. These equations have been widely applied to predict daily energy demands and assess ecosystem carrying capacity \cite{nagy2001food,brown2004toward}.

In cetaceans, body condition and mass have often been estimated from animals obtained through whaling or standings \cite{lockyer1981growth,fortune2012growth}. Body condition is usually inferred from traits such as blubber thickness, lipid composition, and carcass analysis \cite{castrillon2020evaluating}. Although these approaches have yielded valuable insights, they are restricted to dead or captured animals, limiting their usefulness for monitoring living populations. These limitations emphasize the need for reliable, non-invasive methods to evaluate condition and mass in free-ranging cetaceans.
\begin{figure}[t]
    \centering
    \includegraphics[width=0.49\linewidth]{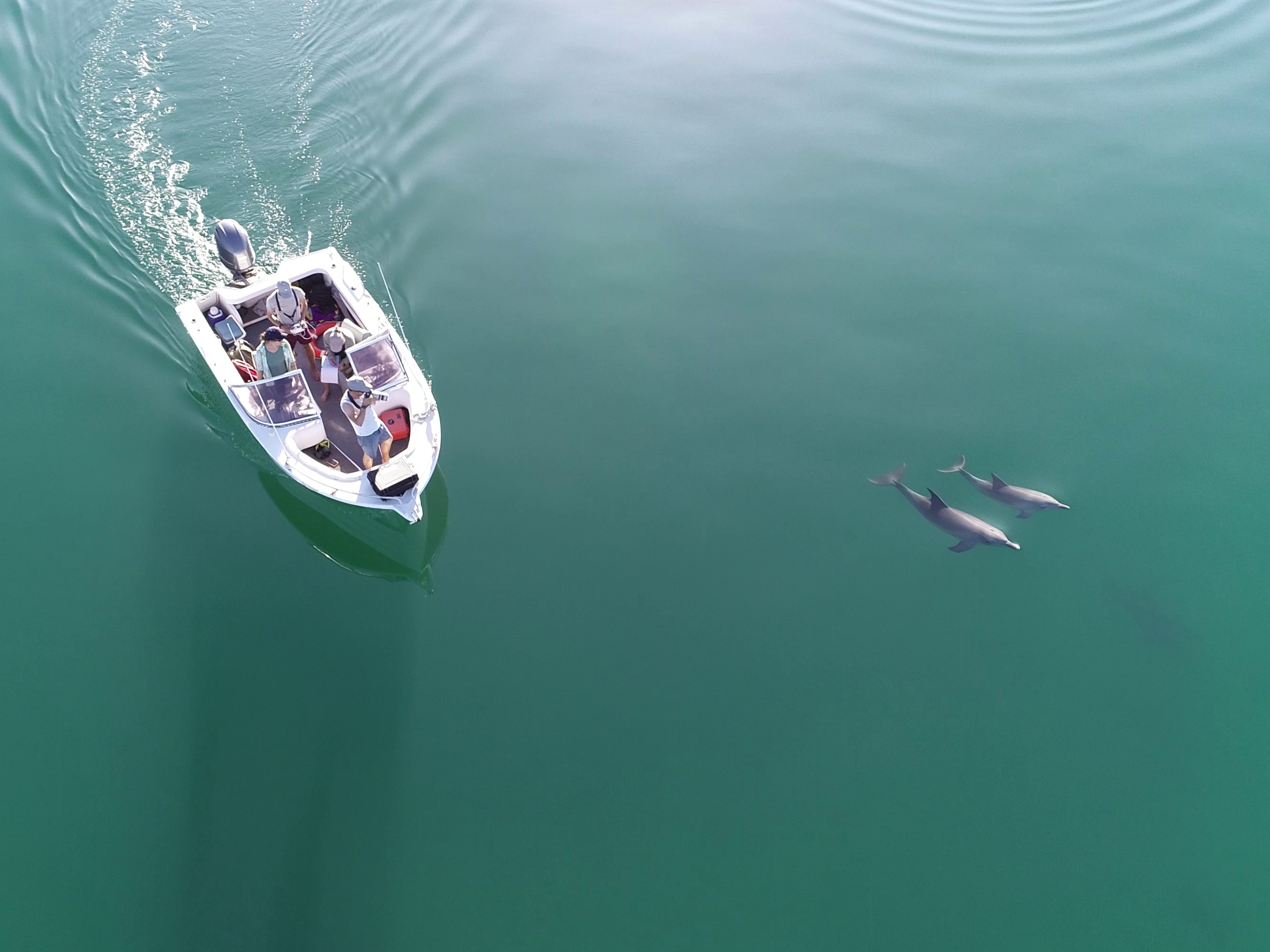}
    \includegraphics[width=0.49\linewidth]{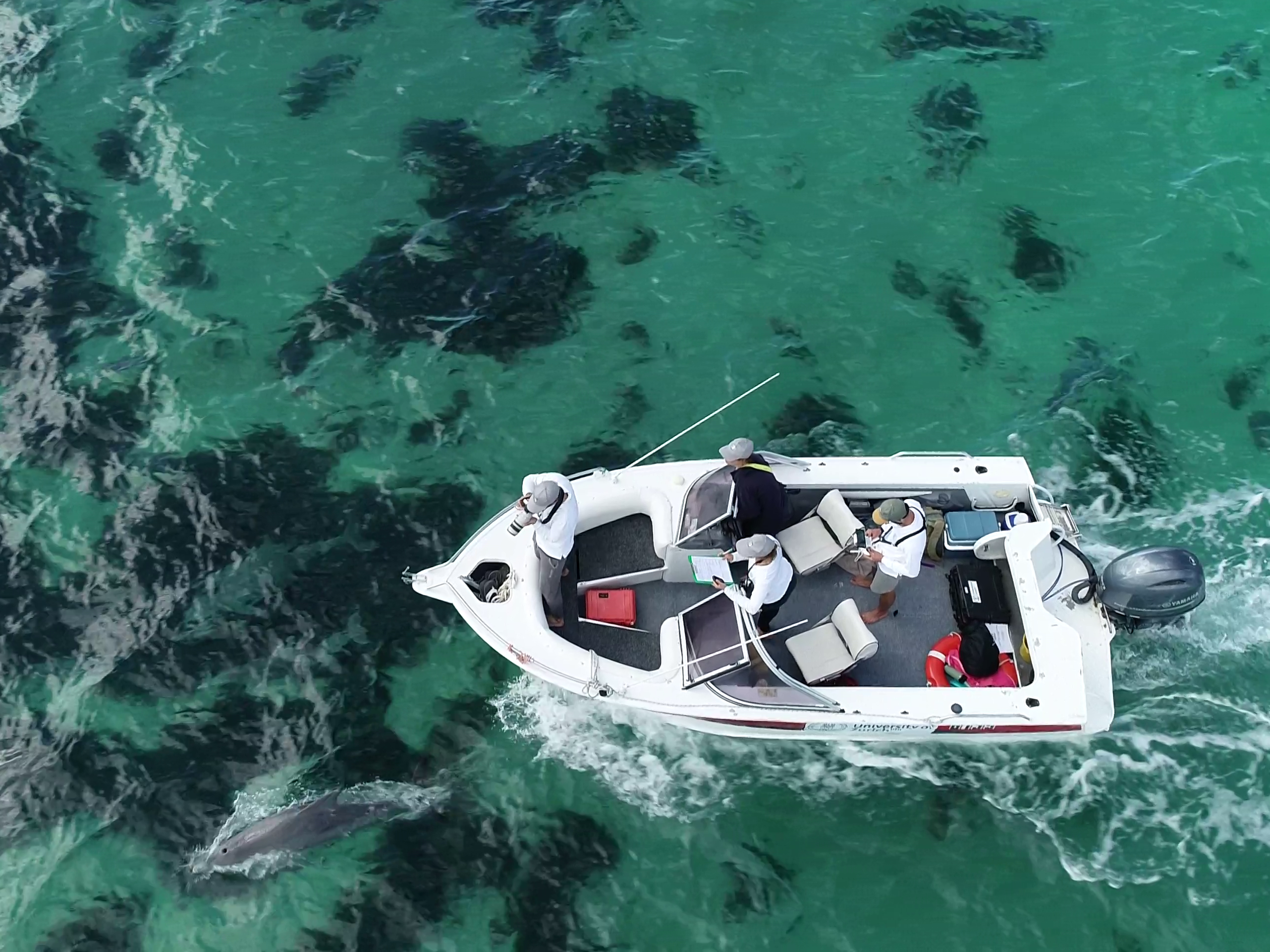}
\caption{\textbf{Data Capture}. We capture footage of dolphins in Shark Bay, Western Australia, where the clear waters allow full-body visibility from drones.}
\label{fig:drone}
\end{figure}

Unoccupied Aerial Vehicle (UAV) photogrammetry provides a powerful, non-invasive approach for remotely estimating body condition and mass \cite{watts2010small, whitehead2014remote, linchant2015unmanned}. While previous studies have used 3D models to estimate mass and condition (e.g., \cite{christiansen2019estimating,glarou2023estimating,irschick2021creation}), these approaches typically do not capture the full 3D pose and shape of freely moving animals. Since body condition and mass are volumetric, reconstructing the metric 3D shape and motion of individuals could improve assessment accuracy.

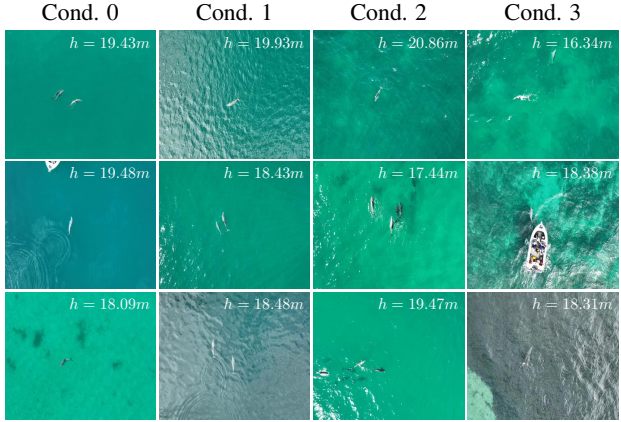
\begin{figure}[t]
    \centering
    \begin{tikzpicture}[image/.style = {inner sep=0pt, outer sep=0pt}, node distance = 0.5mm and 0.5mm] 
        \node[image] (elm) {\includegraphics[width=.24\linewidth, trim=200 50 200 50, clip]{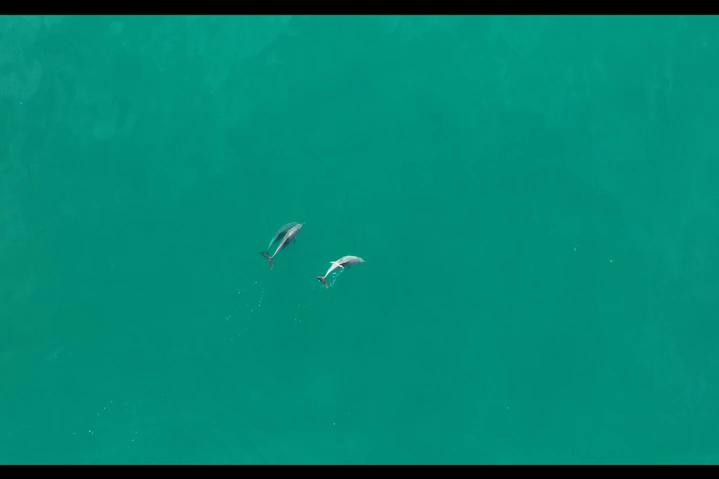}};
        \node[image, right=of elm] (buc) {\includegraphics[width=.24\linewidth, trim=200 50 200 50, clip]{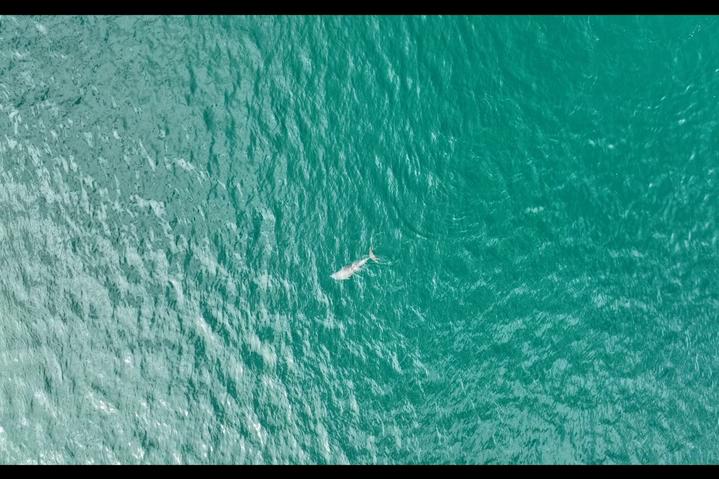}};
        \node[image, right=of buc] (poo) {\includegraphics[width=.24\linewidth, trim=200 50 200 50, clip]{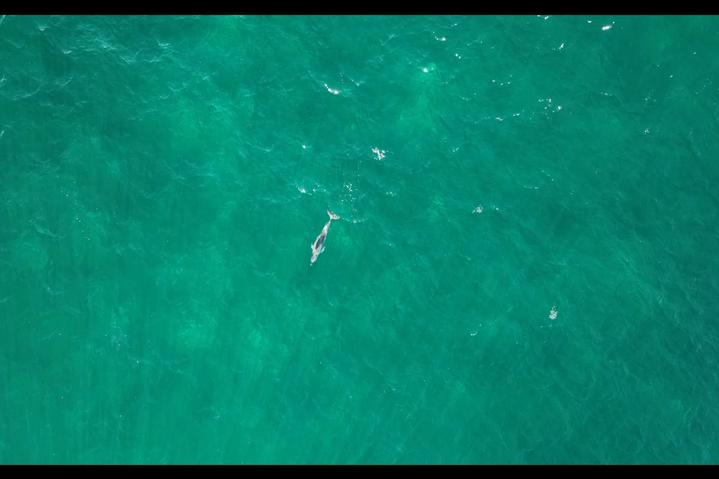}};
        \node[image, right=of poo] (lat) {\includegraphics[width=.24\linewidth, trim=200 50 200 50, clip]{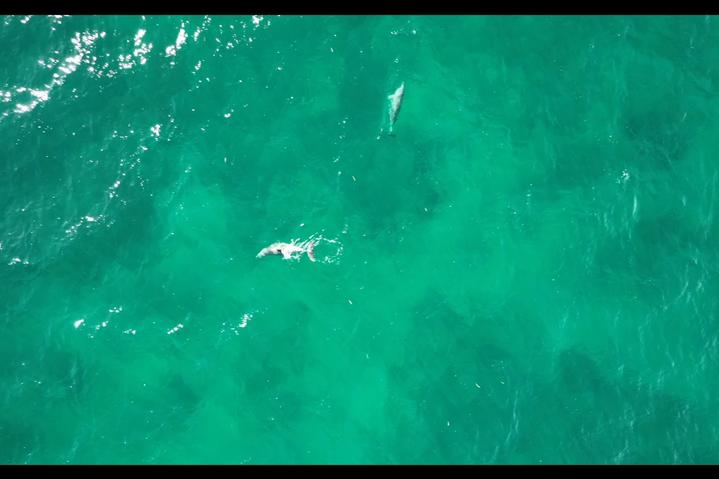}};
        
        \node[image, below=of elm] (hea) {\includegraphics[width=.24\linewidth, trim=200 50 200 50, clip]{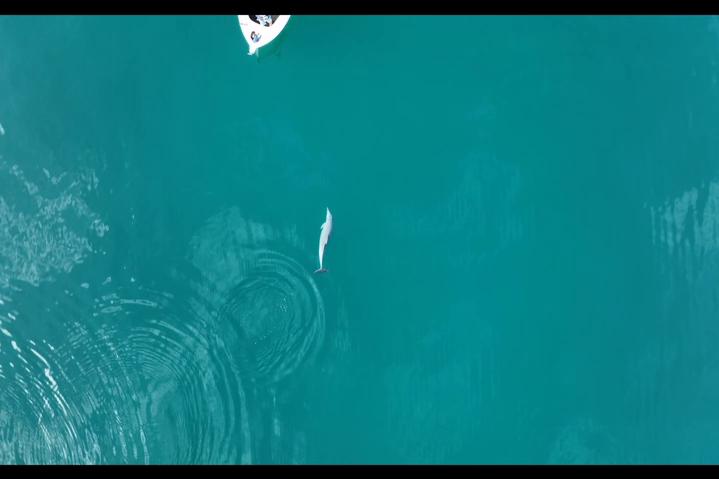}};
        \node[image, right=of hea] (esk) {\includegraphics[width=.24\linewidth, trim=200 50 200 50, clip]{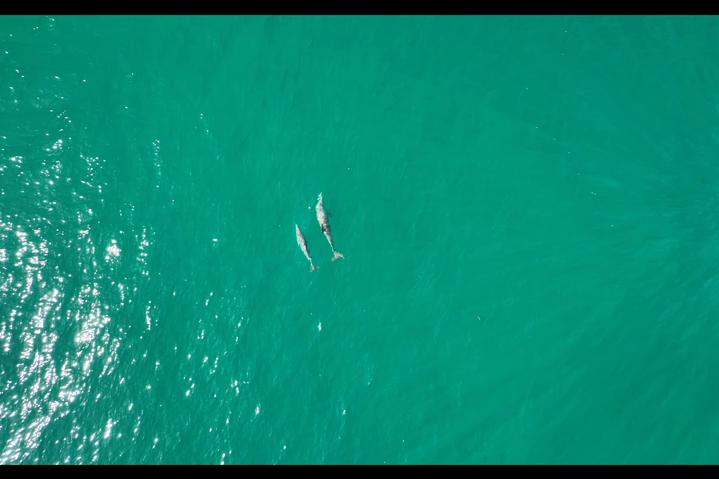}};
        \node[image, right=of esk] (tam) {\includegraphics[width=.24\linewidth, trim=200 50 200 50, clip]{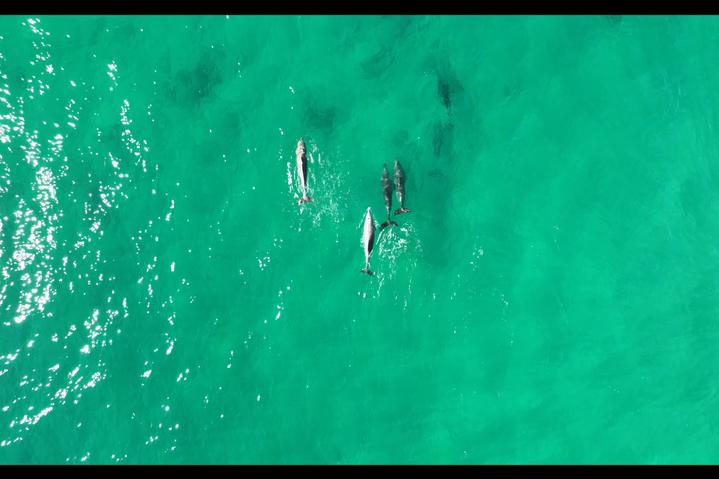}};
        \node[image, right=of tam] (snr) {\includegraphics[width=.24\linewidth, trim=200 50 200 50, clip]{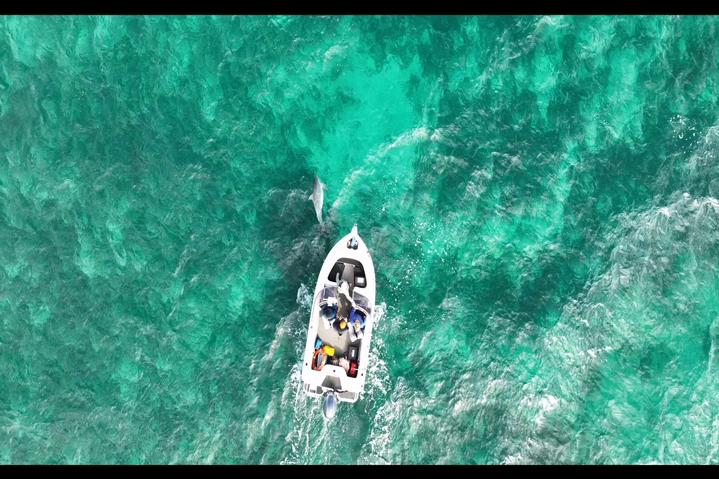}};
        
        \node[image, below=of hea] (slk) {\includegraphics[width=.24\linewidth, trim=200 50 200 50, clip]{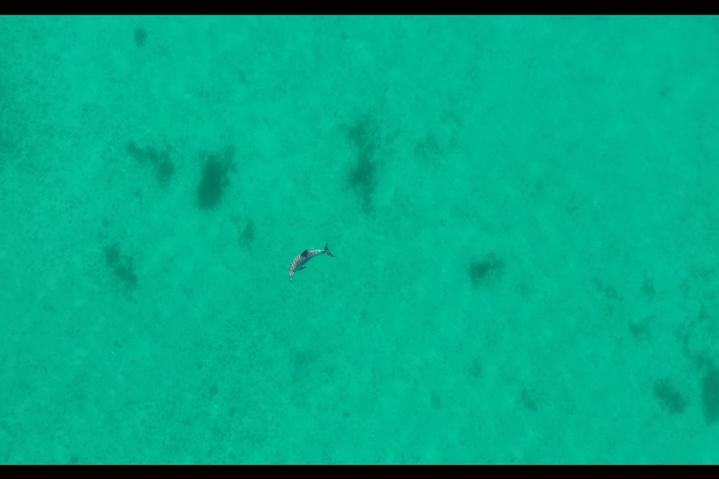}};
        \node[image, right=of slk] (gui) {\includegraphics[width=.24\linewidth, trim=200 50 200 50, clip]{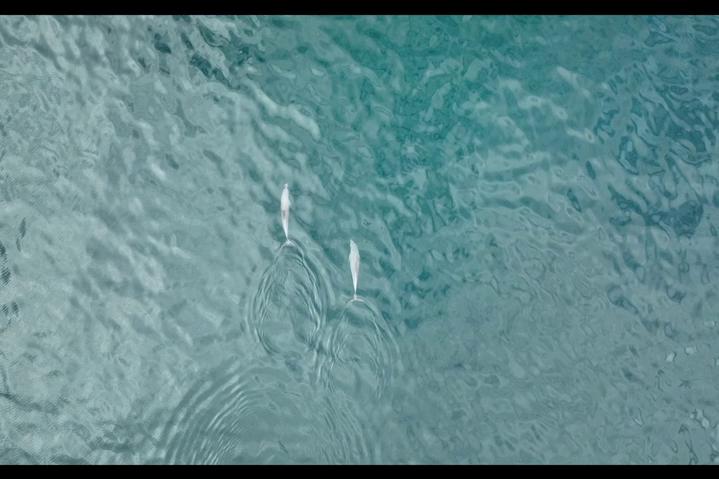}};
        \node[image, right=of gui] (win) {\includegraphics[width=.24\linewidth, trim=200 50 200 50, clip]{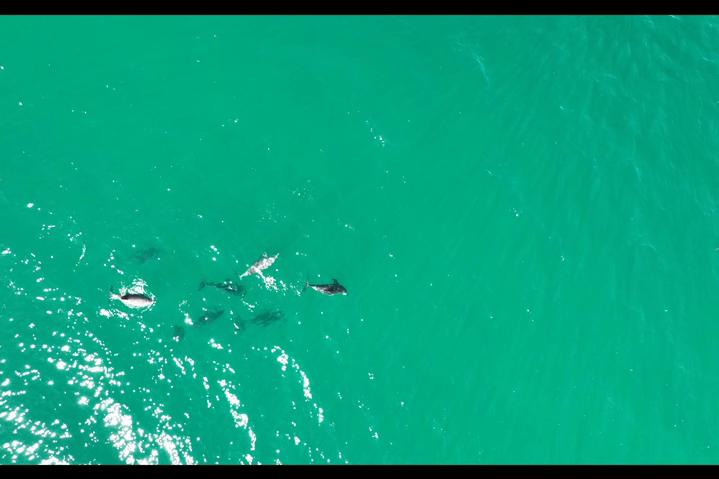}};
        \node[image, right=of win] (wim) {\includegraphics[width=.24\linewidth, trim=200 50 200 50, clip]{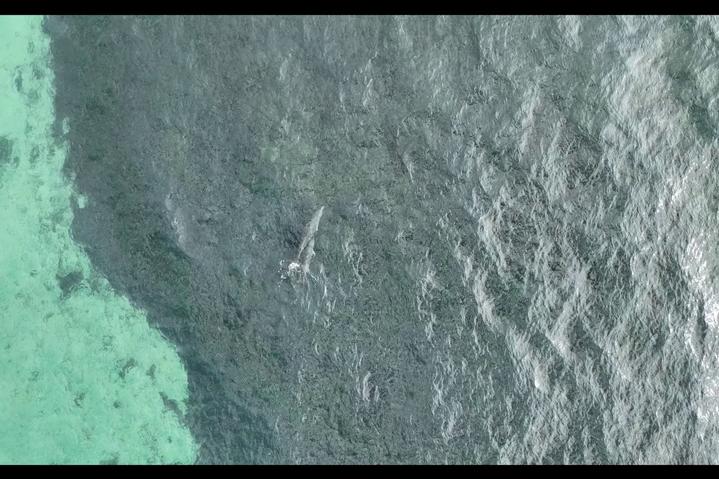}};
        
        \node[anchor=center, scale=0.9] at ([yshift=2.5mm]elm.north) {Cond. 0};
        \node[anchor=center, scale=0.9] at ([yshift=2.5mm]buc.north) {Cond. 1};
        \node[anchor=center, scale=0.9] at ([yshift=2.5mm]poo.north) {Cond. 2};
        \node[anchor=center, scale=0.9] at ([yshift=2.5mm]lat.north) {Cond. 3};

        \node[anchor=north east, scale=0.6] at (elm.north east) {\color{White}$h=19.43m$};
        \node[anchor=north east, scale=0.6] at (buc.north east) {\color{White}$h=19.93m$};
        \node[anchor=north east, scale=0.6] at (poo.north east) {\color{White}$h=20.86m$};
        \node[anchor=north east, scale=0.6] at (lat.north east) {\color{White}$h=16.34m$};
        
        \node[anchor=north east, scale=0.6] at (hea.north east) {\color{White}$h=19.48m$};
        \node[anchor=north east, scale=0.6] at (esk.north east) {\color{White}$h=18.43m$};
        \node[anchor=north east, scale=0.6] at (tam.north east) {\color{White}$h=17.44m$};
        \node[anchor=north east, scale=0.6] at (snr.north east) {\color{White}$h=18.38m$};
        
        \node[anchor=north east, scale=0.6] at (slk.north east) {\color{White}$h=18.09m$};
        \node[anchor=north east, scale=0.6] at (gui.north east) {\color{White}$h=18.48m$};
        \node[anchor=north east, scale=0.6] at (win.north east) {\color{White}$h=19.47m$};
        \node[anchor=north east, scale=0.6] at (wim.north east) {\color{White}$h=18.31m$};
        
    \end{tikzpicture}
\caption{\textbf{Data Collection}. We show examples of frames captured under different Beaufort conditions (see \Cref{sec:data}).}
\label{fig:conditions}
\end{figure}

Motivated by this, we tackle the challenge of estimating the metric 3D shape and motion of dolphins from drone-shot videos captured in the wild, and then use the reconstruction to estimate mass and volume.
Reconstructing 3D animal pose and shape from images and video is complex, due to appearance variation, non-rigid deformations, and occlusion. While significant progress has been made for terrestrial animals, marine species remain largely understudied.
Dolphins, as highly agile aquatic mammals, add further complexity through dynamic, unconstrained motion, water-induced occlusion, and the scarcity of large-scale annotated datasets.

To address monocular ambiguity and data limitations, we adopt a model-based optimization approach. Specifically, we define a parameterized dolphin model, starting from a realistic template, and optimize shape and 3D pose parameters using an analysis-by-synthesis strategy that includes a water transmission model.
Model-based methods for animal~\cite{Zuffi:CVPR:2017} and human~\cite{loper15smpl1} 3D reconstruction rely on articulated parametric models that define low-dimensional shape deformation spaces learned from registered 3D scans.
To be effective, these models must be expressive enough to represent the full population. The shape space captures correlations among all body vertices in the training set, but limited data can hinder metric reconstruction. For example, if small individuals cannot be generated, size differences may be compensated by altering camera distance.

Since scanning a broad population of dolphins with diverse size and shape variations is unfeasible, we define a 3D parametric model using per-part synthetic deformation spaces, including uniform and anisotropic part scaling, similar to GLoSS~\cite{Zuffi:CVPR:2017}. This breaks whole-body correlations and enables representation of a broader range of shapes.
While prior work has used bone-scaling~\cite{badger2020eccv,biggs2020left}, this is insufficient for our goal of estimating body volume.

In our experiments, we recover 3D shape and motion from drone videos captured under varying sea conditions, and compute mass and volume of the reconstructed dolphins.
To our knowledge, this is the first work to reconstruct 3D dolphins from monocular video for mass and volume estimation, and the first to incorporate a water-occlusion model into animal 3D reconstruction.

\section{Related Work}\label{sec:related}
While significant progress has been made in model-based 3D human pose and shape estimation, extending these methodologies to animals presents unique challenges. These include creating suitable 3D parametric shape models that account for inter- and intra-species variations and capturing large datasets with ground-truth 3D pose and shape annotations.
A significant body of work has focused on quadrupeds~\cite{biggs2020left, bite2023rueegg, BARC:2022, li2024dessie, animer2025, genzoo2025} and birds~\cite{badger2020eccv, Wang_2021_CVPR}, yet less attention has been devoted to estimating the 3D pose and shape of aquatic animals such as dolphins. \citet{dolphins} were the first to address the 3D reconstruction of dolphins from images. \citet{Irschick2021} create 3D models of dolphins using photogrammetry. 
\citet{Zuffi:CVPR:2017} introduced SMAL, the first multi-species model analogous to the SMPL model of the human body~\cite{loper15smpl1}, and more recently VAREN for horses~\cite{Zuffi_2024_CVPR_VAREN}. 
Alternative (model-free) approaches aim to reconstruct animals without a 3D parametric shape model ~\cite{cmrKanazawa18, wu2023dove,yang2021lasr,yang2021viser,BANMO,yang2023ppr,wu2023magicpony,yao2022lassie,yao2023hi-lassie,yao2023artic3d}. While these methods are theoretically more general, they often produce unsatisfactory results, require multi-view data, or depend on category-based datasets for learning shape.

\section{Method}\label{sec:method}

We solve our problem as an instance of single-view inverse rendering. Our pipeline requires a single drone-shot video of a dolphin with $T$ frames, labeled with the altitude values of the camera $\{h_i\}_{i=1}^{T}$. The two core parts of our method are 1) a SMPL-like~\cite{loper15smpl1,loper15smpl2} dolphin template model, supporting differentiable skinning and part scaling, and 2) the DIB-R~\cite{chen2019dibr,chen2021dibrpp} differentiable mesh rendering algorithm.

\subsection{Data Collection}\label{sec:data}

In 2024, we collected aerial videos of Indo-Pacific bottlenose dolphins (\textit{Tursiops aduncus}) in Shark Bay, Western Australia, using a DJI Mavic 3 multirotor UAV. The drone was equipped with a 20 MP CMOS four-thirds Hasselblad camera (24 mm f/2.8–f/11 lens) and an SF11/C laser range finder (LightWare LiDAR) to measure altitude above the animals. We operated the drone at altitudes of 15–25 m with the camera angled vertically at -90° and recorded video at 5.1K@50fps.
UAV flights were conducted under favorable weather, restricted to Beaufort 0–3 conditions. At Beaufort 0, the water was mirror-like; Beaufort 1 featured slight ripples without foam crests; Beaufort 2 had small wavelets with occasional crests; and Beaufort 3 introduced short waves with scattered whitecaps, though visibility remained sufficient (see~\Cref{fig:conditions}).
In 2023, we used the same UAV setup to collect morphometric data on 11 T. aduncus dolphins under human care at Sea World Australia. We recorded both dorsal and lateral views as dolphins remained stationary in shallow water, assisted by trainers. For lateral recordings, trainers guided dolphins to rotate their bodies, allowing imaging from either side.
From dorsal images, we measured body length (BL; rostrum tip to tail notch) and body widths (W) at 5 intervals along the body axis \cite{christiansen2016noninvasive, christiansen2018maternal}. Lateral images were used to determine body height (H; dorso-ventral distance) at the same positions \cite{christiansen2019estimating}. In total, we obtained 19 W and H measurements, analyzed using MorphoMetrix software \cite{torres2020morphometrix}.

\subsection{3D Reconstruction}

\subsubsection{Preprocessing}

Inferring 3D information from single-view images is a notoriously hard problem. In order to make it feasible without simplifying assumptions, we need to inject priors into the model. First, we segment the target dolphin in the input videos using the SAM-2 model~\cite{ravi2025sam}. Thanks to its excellent tracking abilities, SAM-2 provides reasonable masks in most cases, even when the dolphins would dive deep and gradually disappear. Then, we employ this information to initialize the position of the dolphin for all frames: by exploiting the fact that all images are perpendicular drone shots with known altitude, we can map the center of the dolphin geometry in world space to the center of the masked pixels in image space by inverting the camera transform.

\begin{figure}[t]

    \centering
    \begin{tikzpicture}[image/.style = {inner sep=0pt, outer sep=0pt}, node distance = 0mm and 0.5mm] 

        \node[image] (depth-0) {{\includegraphics[width=.24\linewidth, trim=190 40 200 70, clip]{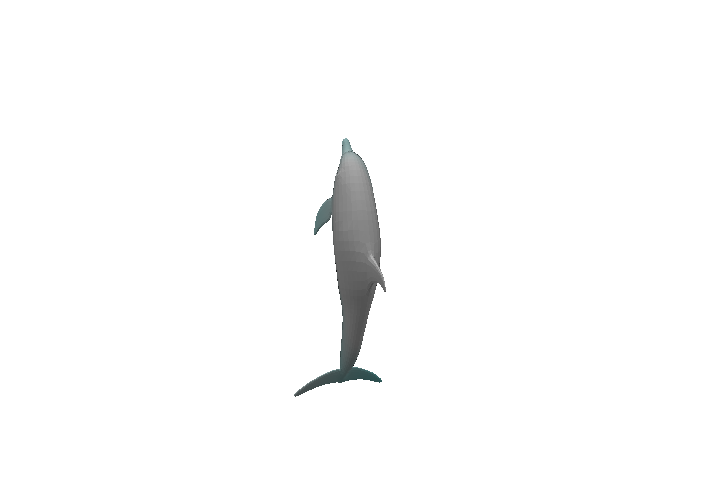}}};
        \node[image, right=of depth-0] (depth-1) {{\includegraphics[width=.24\linewidth, trim=190 40 200 70, clip]{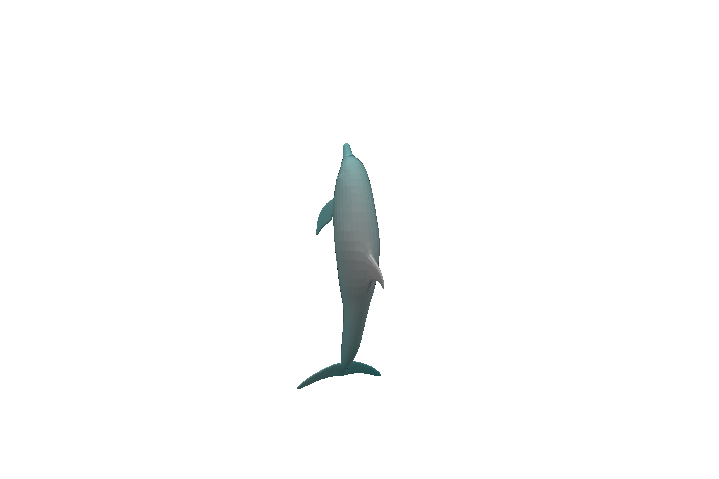}}};
        \node[image, right=of depth-1] (depth-2) {{\includegraphics[width=.24\linewidth, trim=190 40 200 70, clip]{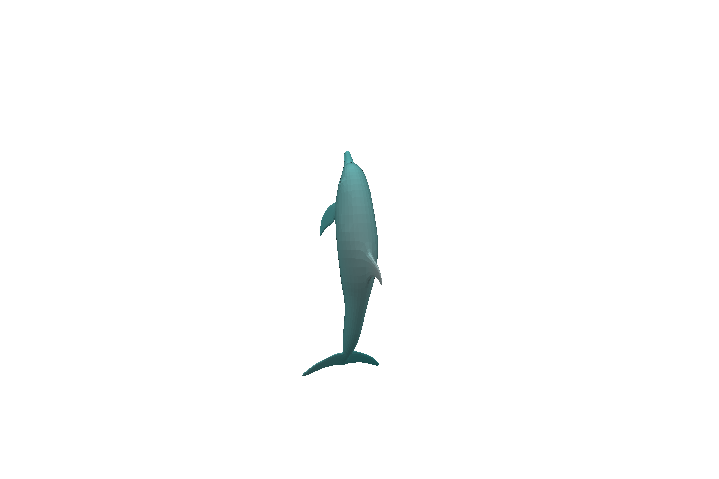}}};
        \node[image, right=of depth-2] (depth-3) {{\includegraphics[width=.24\linewidth, trim=190 40 200 70, clip]{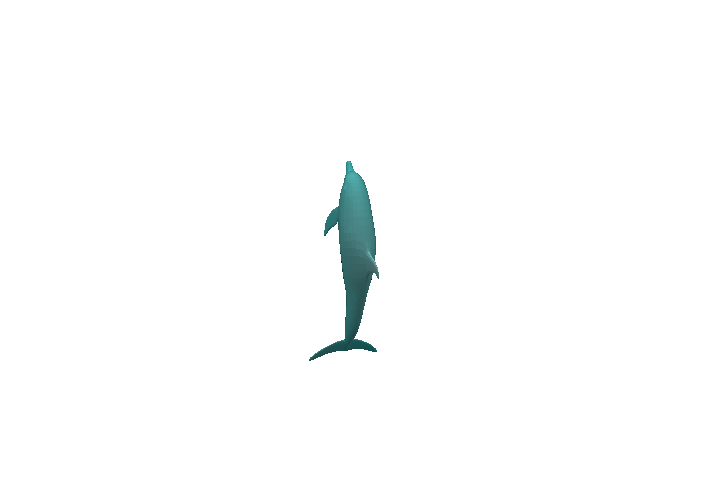}}};

        \draw[-{Stealth[scale=1.5]}] (depth-0.north) -- (depth-3.north) node [midway, above, scale=0.9] {Depth};
        \fill (depth-0.north)  circle[radius=2pt];

        \node[fill=red, anchor=south, scale=0.6, text width=.4\linewidth, align=center,rounded corners=.1cm] at ([xshift=80pt,yshift=-10pt]depth-0.north) {\color{White}Water filter color};
        
    \end{tikzpicture}
\caption{\textbf{Water filtering.} We show the effect of our water filter, with a fixed color (bottom right), by varying the distance of the dolphin mesh from the water plane.}
\end{figure}
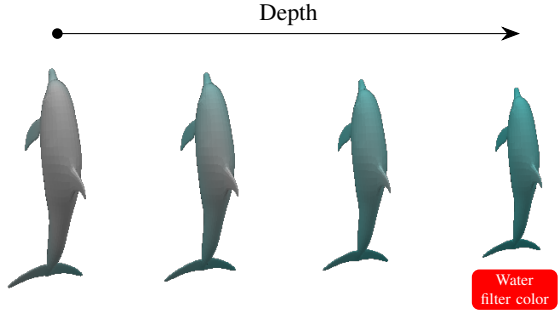

\subsubsection{Model}

We model various features of the scene depicted in the input videos, in order to ensure the most accurate reconstruction of the dolphin geometry.

\begin{figure}[t]
\includegraphics[width=1\linewidth]{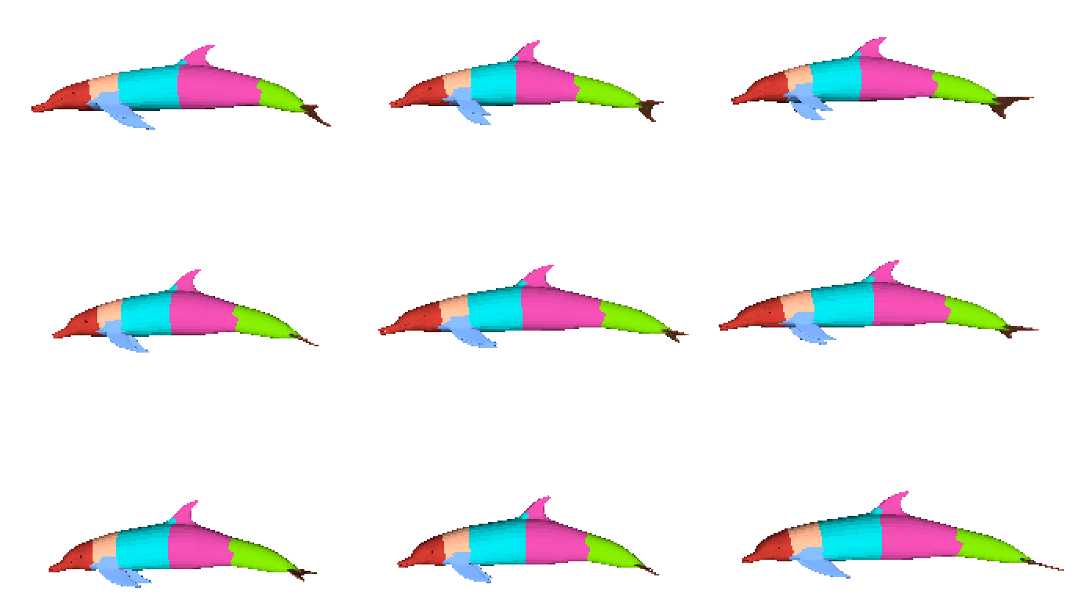}
\caption{\textbf{Samples}. We show samples from the dolphin model; note how we can create different shapes my deforming the body parts.}
\label{fig:samples}
\end{figure}

\paragraph{Parametric Dolphin Model}

The parametric 3D dolphin model follows a representation similar to 3D parametric shape models of humans and animals, computing the vertices of a triangular mesh given pose and shape parameters. 
The model is a function $M(\beta, \vartheta)$ of shape $\beta$ and pose $\vartheta$. $\beta$ is the matrix of shape variables, while $\vartheta \in \mathbb{R}^{3\times N}$ 
is the relative rotation, expressed with Rodrigues vectors, of the $N$ joints in the kinematic tree.

Differently from previous models, in the dolphin model the shape variables are not a vector, but a matrix, as a vector of shape variables is defined for each body part. 
The model computes the coordinates of the vertices in the template pose for each body part as:
\begin{equation}
\textbf{v}_{\text{shape}, I_i}(\beta_i) = \textbf{v}_{t, I_i} + B_{I_i} \beta_i,\\
\label{eq:sal}
\end{equation}
where $\textbf{v}_{t}$ represents the vertices of the dolphin template, $B$ is a matrix of deformation vectors,
$\beta_i$ is the shape vector corresponding to the part $i$. $I_i$ is the set of vertex indices that belong to the part $i$. 
Given vertices of different parts are independently modified, the resulting model is not smoothly connected at the part skin interfaces.
Joint displacements are computed to connect each part to its parent in the kinematic tree.
The matrix of shape components $B$ is defined as the synthetic shape space of the GLoSS model~\cite{Zuffi:CVPR:2017}. Samples from the model are shown in Fig.~\ref{fig:samples}.

We group the $N$ joints of our model into $M\leq N$ parts, allowing to account for symmetries in motion and scale (\eg, fins, flukes). These are controlled by the pose parameters $\bm{\theta}\in\mathbb{R}^{T\times M\times 3}$ and the (non-time dependent) shape parameters $\bm{\beta}\in\mathbb{R}^{M\times 4}$. The first group/joint represents the root bone, \ie, it controls the global orientation of the dolphin. The position of the dolphin, on the other hand, is modeled as a per-frame translation of the root bone $\bm{P}\in\mathbb{R}^{T\times 3}$.

\paragraph{Appearance Modeling}

Our dolphin model comes with UV maps, which allows us to represent its appearance via a color texture $I_{\text{albedo}}\in [0;1]^{H_a\times W_a}$. We made it grayscale to prevent water effects from bleeding into the dolphin appearance during optimization. We then modeled such effects separately by introducing a novel water filter. After obtaining a render $C\in[0;1]^{H_r\times W_r\times 3}$ of the scene, we post-process it as: 
\begin{equation}
\hat{C}_{i,j} = C_{i,j} \circ \exp(\min(d_{i,j}, 0) \cdot F_{\text{water}}),
\end{equation}
where $F_{\text{water}}\in[0;1]^3$ is a parameter modeling which color frequencies are filtered by a volume of water, and $d_{i,j}$ is the (negative) depth of the dolphin at pixel $(i,j)$. This information is available via the camera altitude, which allows us to assume that the water plane lies at 0 along the up axis of the scene.
Lastly, illumination effects are captured by a simple spherical Gaussians~\cite{sphericalgaussians} model with parameters $S$. 

\begin{figure}[t]
    \centering
    \begin{tikzpicture}[image/.style = {inner sep=0pt, outer sep=0pt}, node distance = 0mm and 0.5mm] 

        \node[image] (hera-photo) {\includegraphics[width=.22\linewidth, trim=160 100 200 100, clip]{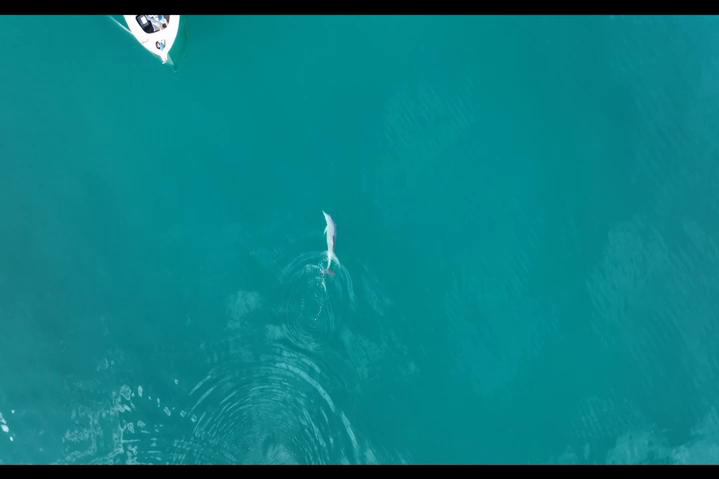}};
        \node[image, right=of hera-photo] (guinness-photo) {\includegraphics[width=.22\linewidth, trim=185 85 175 115, clip]{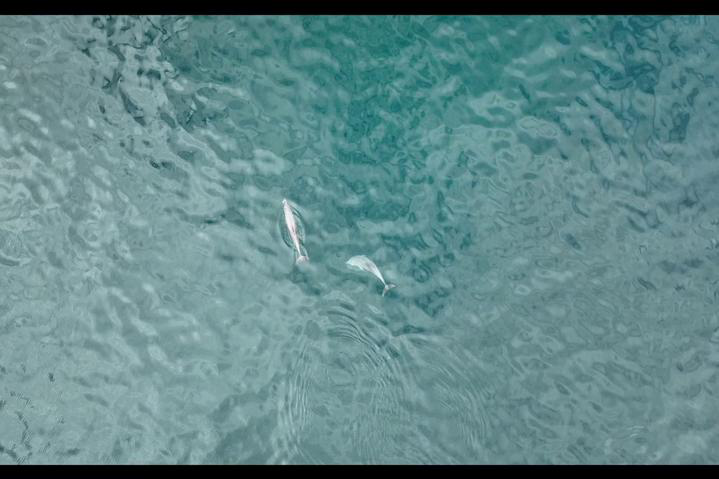}};
        \node[image, right=of guinness-photo] (tam-photo) {\includegraphics[width=.22\linewidth, trim=190 80 170 120, clip]{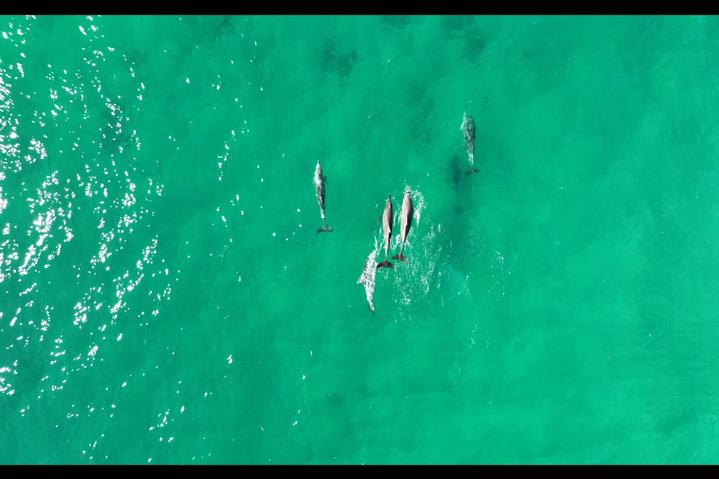}};
        \node[image, right=of tam-photo] (simba-photo) {\includegraphics[width=.22\linewidth, trim=190 130 170 70, clip]{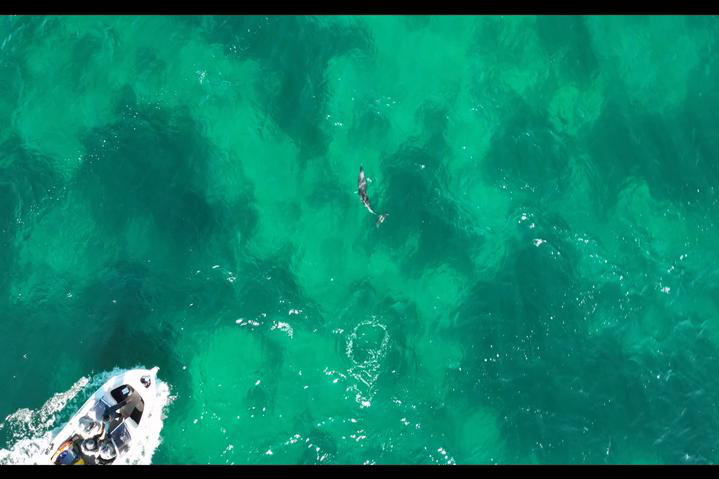}};
        
        \node[image, below=of hera-photo] (hera-mask) {\includegraphics[width=.22\linewidth, trim=160 100 200 100, clip]{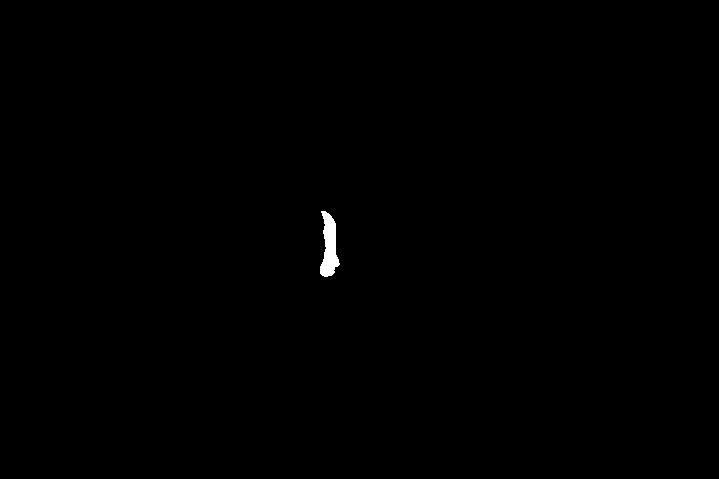}};
        \node[image, right=of hera-mask] (guinness-mask) {\includegraphics[width=.22\linewidth, trim=185 85 175 115, clip]{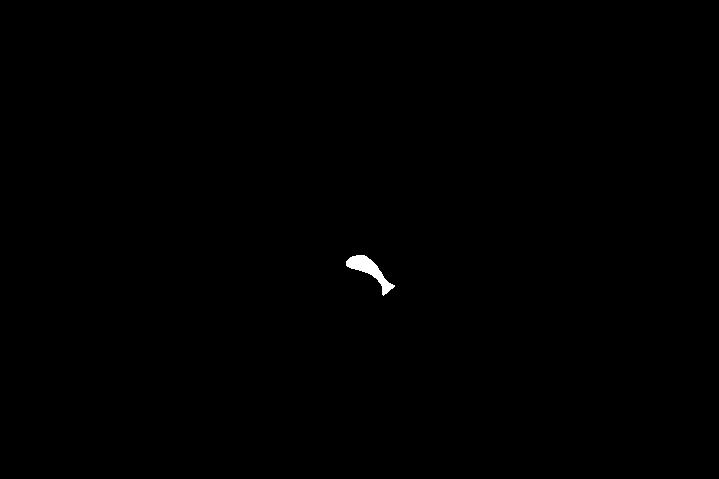}};
        \node[image, right=of guinness-mask] (tam-mask) {\includegraphics[width=.22\linewidth, trim=190 80 170 120, clip]{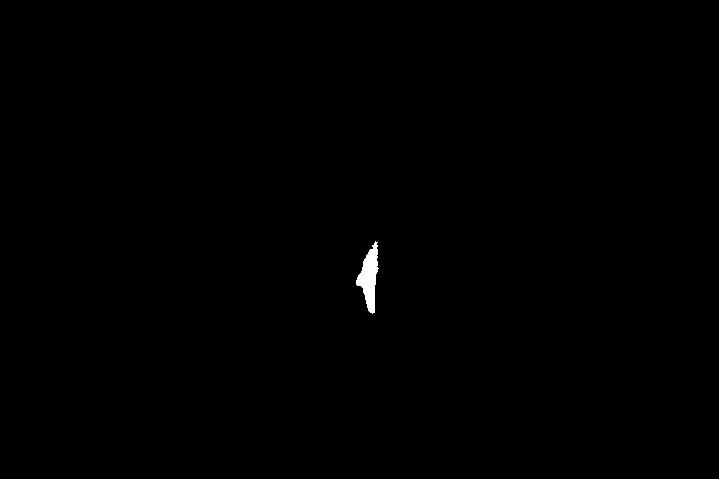}};
        \node[image, right=of tam-mask] (simba-mask) {\includegraphics[width=.22\linewidth, trim=190 130 170 70, clip]{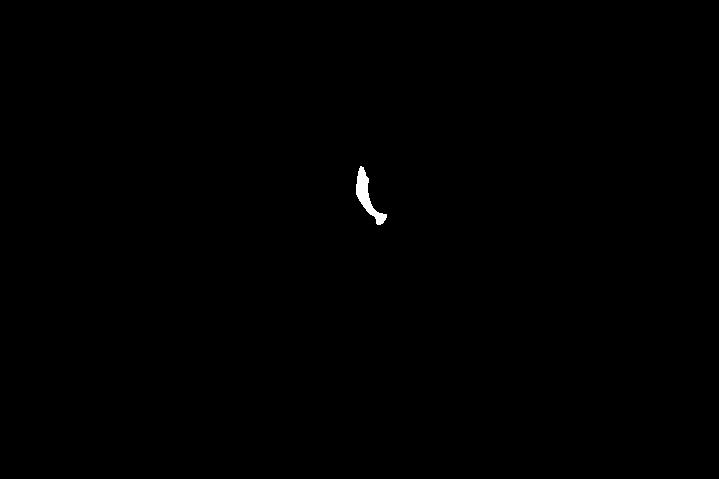}};
        
        \node[image, below=of hera-mask] (hera-render) {\includegraphics[width=.22\linewidth, trim=160 100 200 100, clip]{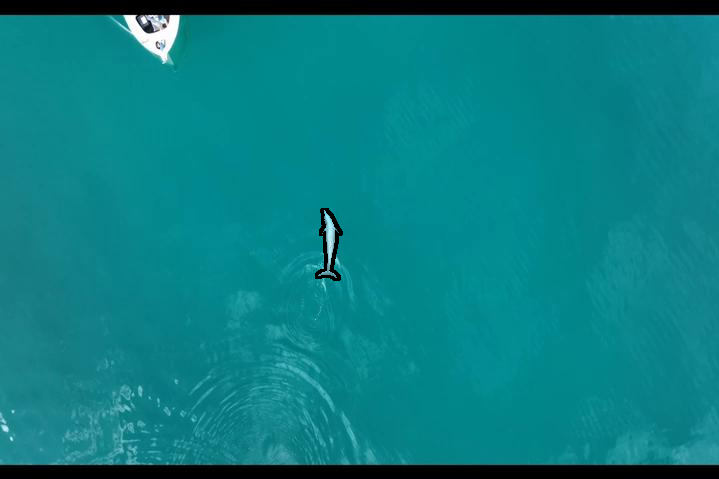}};
        \node[image, right=of hera-render] (guinness-render) {\includegraphics[width=.22\linewidth, trim=185 85 175 115, clip]{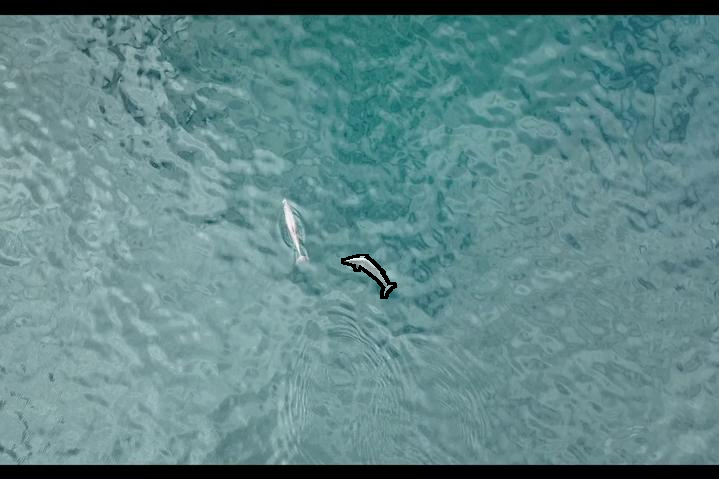}};
        \node[image, right=of guinness-render] (tam-render) {\includegraphics[width=.22\linewidth, trim=190 80 170 120, clip]{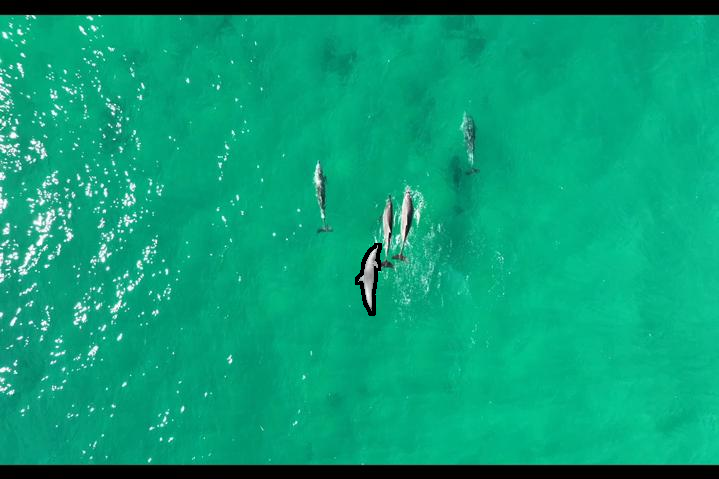}};
        \node[image, right=of tam-render] (simba-render) {\includegraphics[width=.22\linewidth, trim=190 130 170 70, clip]{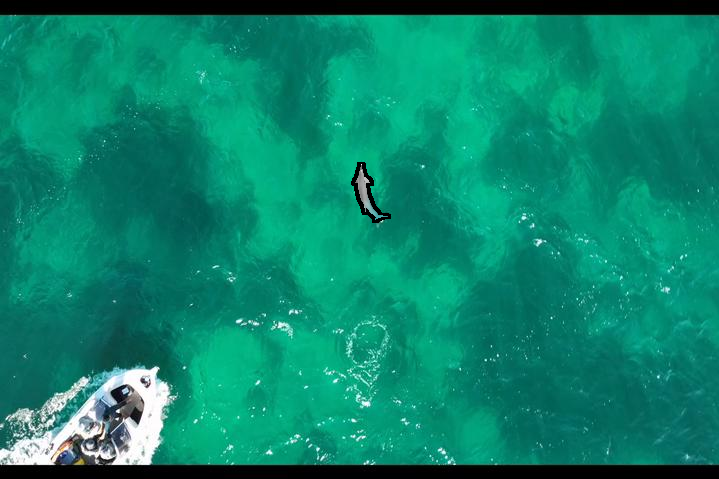}};

        \node[rotate=90, anchor=center, scale=0.9] at ([xshift=-2.5mm]hera-photo.west) {Frame};
        \node[rotate=90, anchor=center, scale=0.9] at ([xshift=-2.5mm]hera-mask.west) {Mask};
        \node[rotate=90, anchor=center, scale=0.9] at ([xshift=-2.5mm]hera-render.west) {Render};
        
        \node[anchor=center, scale=0.9] at ([yshift=2.5mm]hera-photo.north) {HEA};
        \node[anchor=center, scale=0.9] at ([yshift=2.5mm]guinness-photo.north) {GUI};
        \node[anchor=center, scale=0.9] at ([yshift=2.5mm]tam-photo.north) {TAM};
        \node[anchor=center, scale=0.9] at ([yshift=2.5mm]simba-photo.north) {SIM};
        
    \end{tikzpicture}
\caption{\textbf{Image Reconstruction.} For 4 scenes in our dataset, we show (from top to bottom) the original video frame, the SAM-2 segmentation, and the rendering obtained after optimization.}
\label{fig:rendering}
\end{figure}
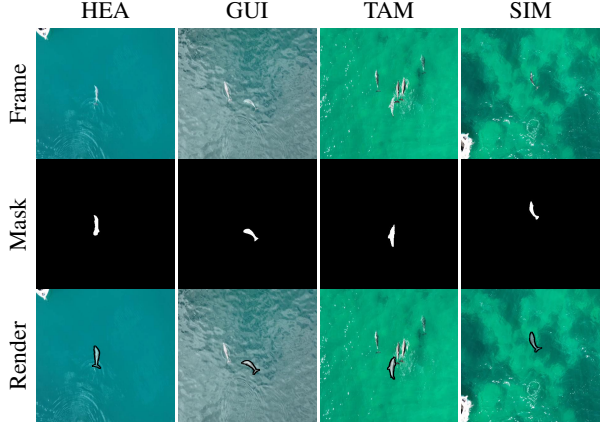

\subsubsection{Optimization}

In the previous section, we introduced the optimization parameters, which we summarize as $\Theta = \{\bm{\theta}, \bm{\beta}, \bm{P}, I_{\text{albedo}}, F_{\text{water}}, S\}$.
Additional to the altitude and pose, we employ the intrinsic properties of the drone camera to ensure correspondence in length units between virtual and real space. Using the sensor width $\gamma$ and focal length $\omega$, the field of view of the virtual camera is defined as $\text{fov} = 2\cdot \tan^{-1}\left(\frac{\gamma}{2\omega}\right)$. The base dolphin model is rescaled so that its initial length its 2.6m, which is the average length for the species of dolphin depicted in our data.
The loss function we use to optimize our scene is defined as the sum of multiple terms, which we describe in the following. The loss function is optimized iteratively via Adam~\cite{Kingma2014AdamAM} steps.

\begin{figure}[t]
    \centering
    \begin{tikzpicture}[image/.style = {inner sep=0pt, outer sep=0pt}, node distance = 0mm and 0.5mm] 

        \node[image] (hera-photo) {\includegraphics[width=.22\linewidth, trim=120 70 120 20, clip]{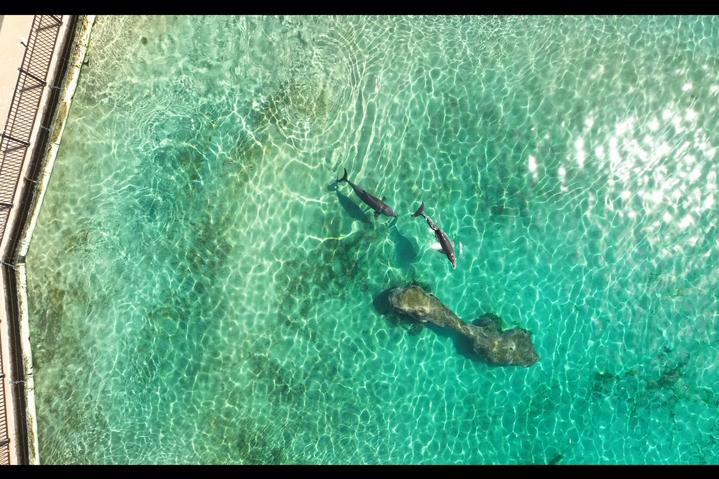}};
        \node[image, right=of hera-photo] (guinness-photo) {\includegraphics[width=.22\linewidth, trim=120 70 120 20, clip]{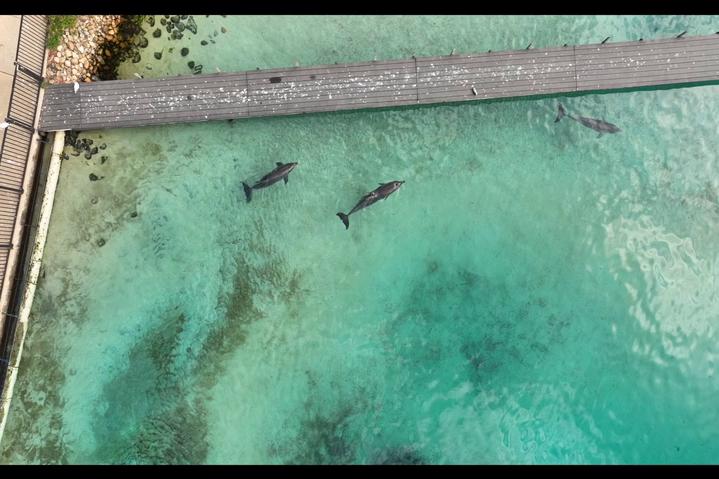}};
        \node[image, right=of guinness-photo] (tam-photo) {\includegraphics[width=.22\linewidth, trim=120 20 120 70, clip]{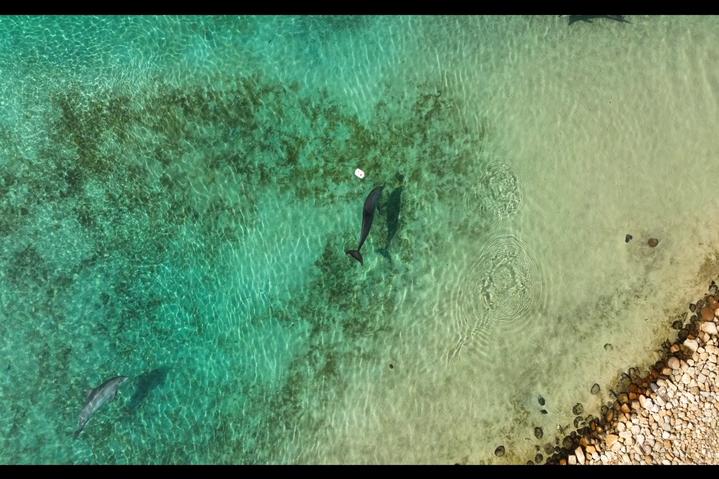}};
        \node[image, right=of tam-photo] (simba-photo) {\includegraphics[width=.22\linewidth, trim=120 70 120 20, clip]{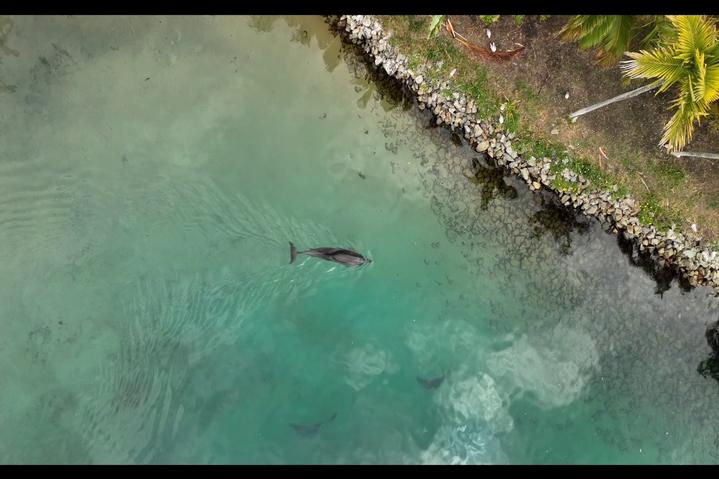}};
        
        \node[image, below=of hera-photo] (hera-mask) {\includegraphics[width=.22\linewidth, trim=120 70 120 20, clip]{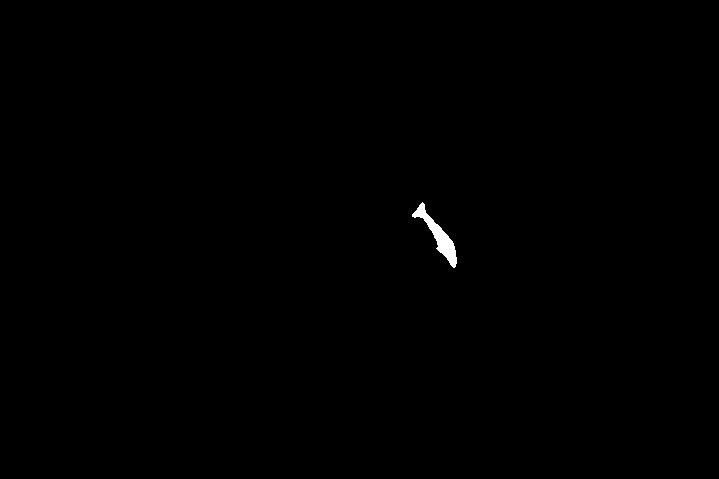}};
        \node[image, right=of hera-mask] (guinness-mask) {\includegraphics[width=.22\linewidth, trim=120 70 120 20, clip]{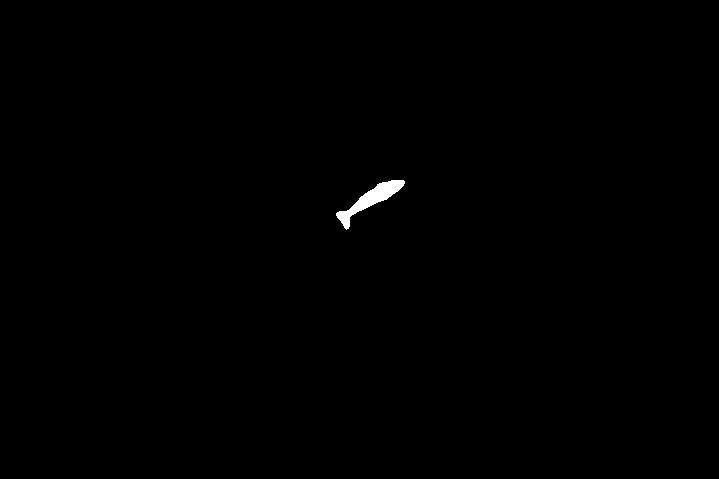}};
        \node[image, right=of guinness-mask] (tam-mask) {\includegraphics[width=.22\linewidth, trim=120 70 120 20, clip]{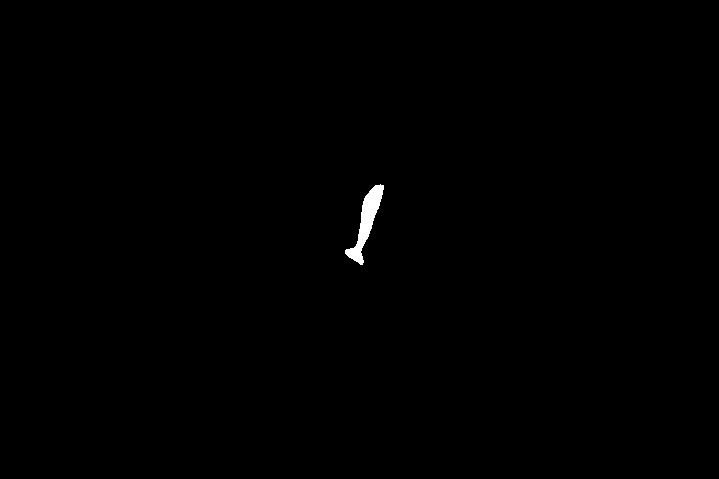}};
        \node[image, right=of tam-mask] (simba-mask) {\includegraphics[width=.22\linewidth, trim=120 70 120 20, clip]{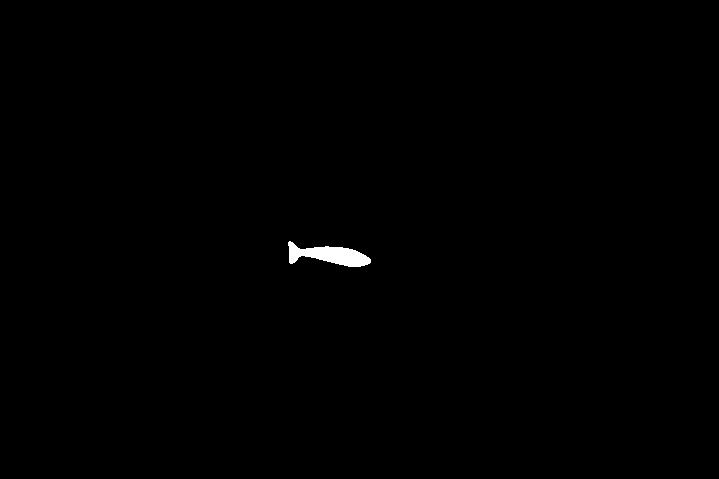}};
        
        \node[image, below=of hera-mask] (hera-render) {\includegraphics[width=.22\linewidth, trim=120 70 120 20, clip]{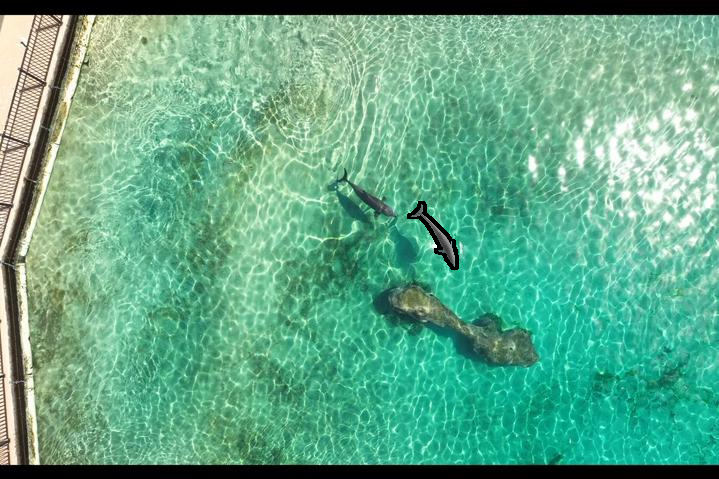}};
        \node[image, right=of hera-render] (guinness-render) {\includegraphics[width=.22\linewidth, trim=120 70 120 20, clip]{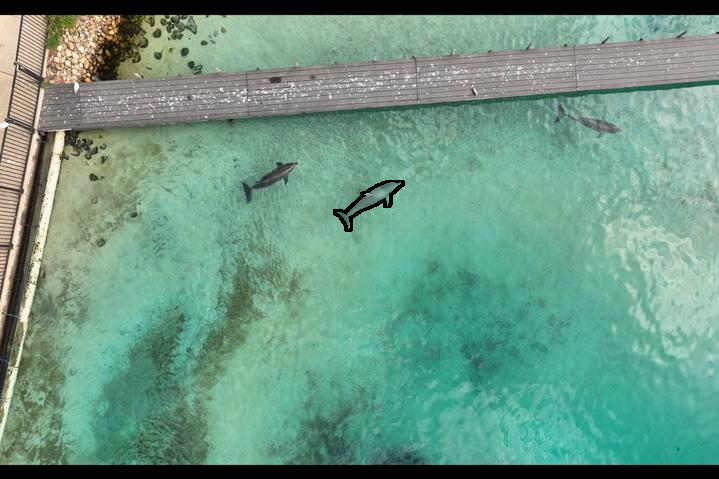}};
        \node[image, right=of guinness-render] (tam-render) {\includegraphics[width=.22\linewidth, trim=120 20 120 70, clip]{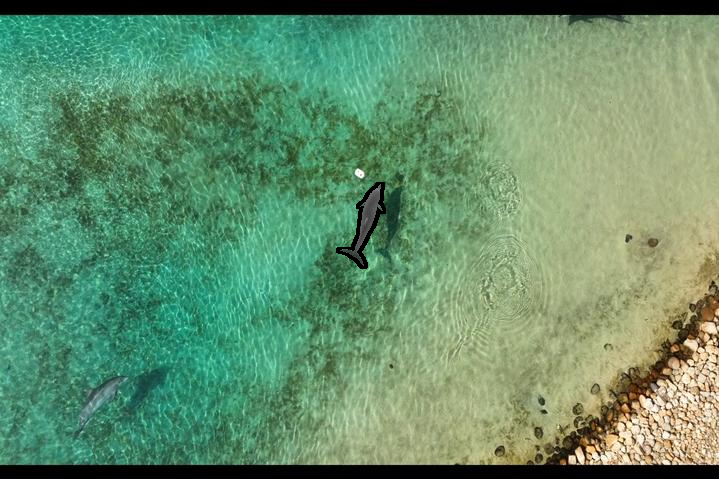}};
        \node[image, right=of tam-render] (simba-render) {\includegraphics[width=.22\linewidth, trim=120 70 120 20, clip]{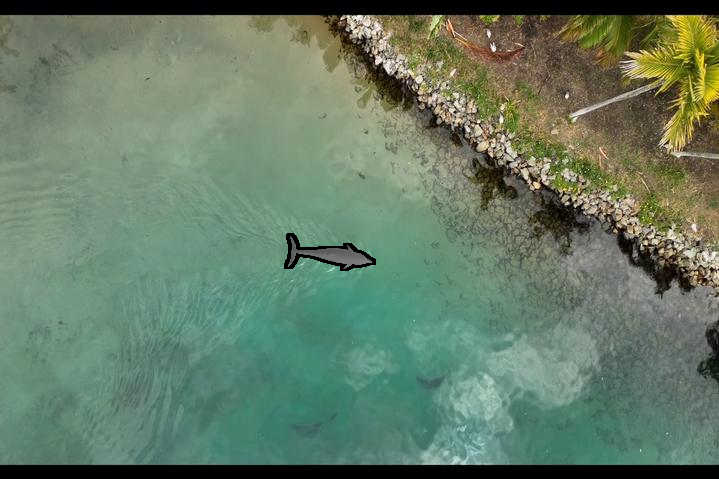}};

        \node[rotate=90, anchor=center, scale=0.9] at ([xshift=-2.5mm]hera-photo.west) {Frame};
        \node[rotate=90, anchor=center, scale=0.9] at ([xshift=-2.5mm]hera-mask.west) {Mask};
        \node[rotate=90, anchor=center, scale=0.9] at ([xshift=-2.5mm]hera-render.west) {Render};
        
        \node[anchor=center, scale=0.9] at ([yshift=2.5mm]hera-photo.north) {Huey (2)};
        \node[anchor=center, scale=0.9] at ([yshift=2.5mm]guinness-photo.north) {Howie};
        \node[anchor=center, scale=0.9] at ([yshift=2.5mm]tam-photo.north) {Tallula};
        \node[anchor=center, scale=0.9] at ([yshift=2.5mm]simba-photo.north) {Scooter};
        
    \end{tikzpicture}
\caption{\textbf{SeaWorld Results.} For 4 SeaWorld specimen, we show (from top to bottom) the original video frame, the SAM-2 segmentation, and the rendering obtained after optimization.}
\label{fig:seaworld}
\end{figure}

\paragraph*{Reconstruction Losses} Our data fidelity terms follow common choices from past literature. Given the complete render of the current scene model $\hat{C}^{(t)}$ at frame $t$, the corresponding ground truth image $I^{(t)}$ and the mask $B\in\{0,1\}^{H_r\times W_r}$ specifying which pixels were mapped to the dolphin mesh during rendering, we compute the photometric loss as:
\begin{equation}
    L_{\text{RGB}} = \sum_{i=1}^{H_r} \sum_{j=1}^{W_r} \left\lVert B_{i,j} \cdot \left(\hat{C}^{(t)}_{i,j} - I^{(t)}_{i,j}\right) \right\rVert_2^2
\end{equation}
Given the output DIB-R soft mask $\hat{B}^{(t)}$ at frame $t$, we optimize the dolphin shape with respect to the SAM-2 segmentation mask $\bar{B}$ using an Intersection-over-Union loss:
\begin{equation}
    L_{\text{mask}} = 1 - \sum_{i=1}^{H_r} \sum_{j=1}^{W_r} \dfrac{\hat{B}^{(t)}_{i,j}\bar{B}^{(t)}_{i,j}}{\left(\hat{B}^{(t)}_{i,j} + \bar{B}^{(t)}_{i,j}\right) - \hat{B}^{(t)}_{i,j}\bar{B}^{(t)}_{i,j}}
\end{equation}

\paragraph*{Distribution Terms} To avoid obtaining degenerate behaviour with out-of-distribution values in the pose or scale parameters, we regularize both to have minimal L2 norm (\ie, values close to zero) via the $L_{\text{pose}}$ and $L_{\text{scale}}$ terms.

\paragraph*{Scale Regularizers} Given that the ``depth'' dimension of the dolphin (corresponding in our model to the $y$ axis) is usually unobserved in the data, we want to avoid its scale value being completely independent from the dolphin's size in the other (observed) dimensions. To this end, we penalize differences in scale for distinct axes of the same group:
\begin{equation}
    L_{\text{scale}}^{\text{axes}} = \sum_{k=1}^{M} \lVert \bm{\beta}_{k,x} - \bm{\beta}_{k,y} \rVert_2^2 + \lVert \bm{\beta}_{k,y} - \bm{\beta}_{k,z} \rVert_2^2
\end{equation}
Then, to avoid large variations of scale in adjacent model parts, we employ the model skeleton's structure to penalize differences in the scale of the $k$-th bone group and its parent $p(k)$ (which is undefined for the root bone, thus we ignore it):
\begin{equation}
    L_{\text{scale}}^{\text{smooth}} = \sum_{k=2}^{M} \left\lVert \bm{\beta}_{k} - \bm{\beta}_{p(k)} \right\rVert_2^2
\end{equation}

\paragraph*{Smoothness Terms} We regularize the variation over time of per-frame parameters (dolphin position, orientation and joints pose) by minimizing the MSE between their values at adjacent frames. In practice, fitting the terms $L_{\text{smooth}}^{\text{pos}}$, $L_{\text{smooth}}^{\text{rot}}$, and $L_{\text{smooth}}^{\text{joints}}$ along with our reconstruction losses allows us to obtain smooth animations.

\begin{figure}[t]
    \centering
    \begin{tikzpicture}[image/.style = {inner sep=0pt, outer sep=0pt}, node distance = 0.5mm and 0.5mm] 

        \node[image] (hera-render) {\includegraphics[height=.175\linewidth, trim=190 95 210 145, clip]{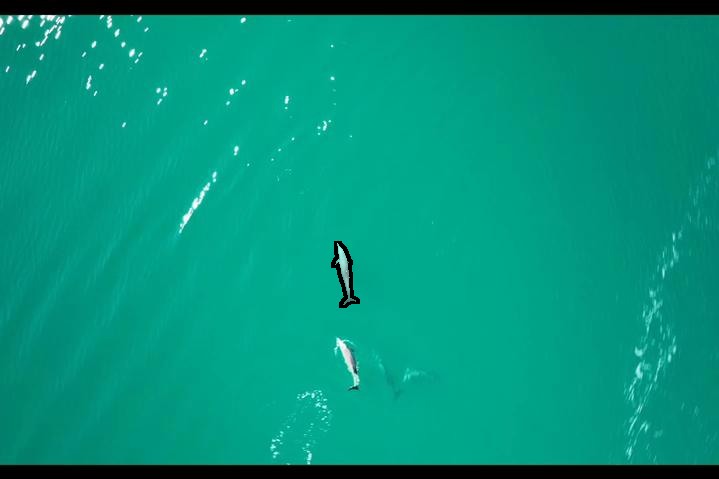}};
        \node[image, right=of hera-render] (hera-mesh) {\includegraphics[width=.35\linewidth, trim=50 100 50 90, clip]{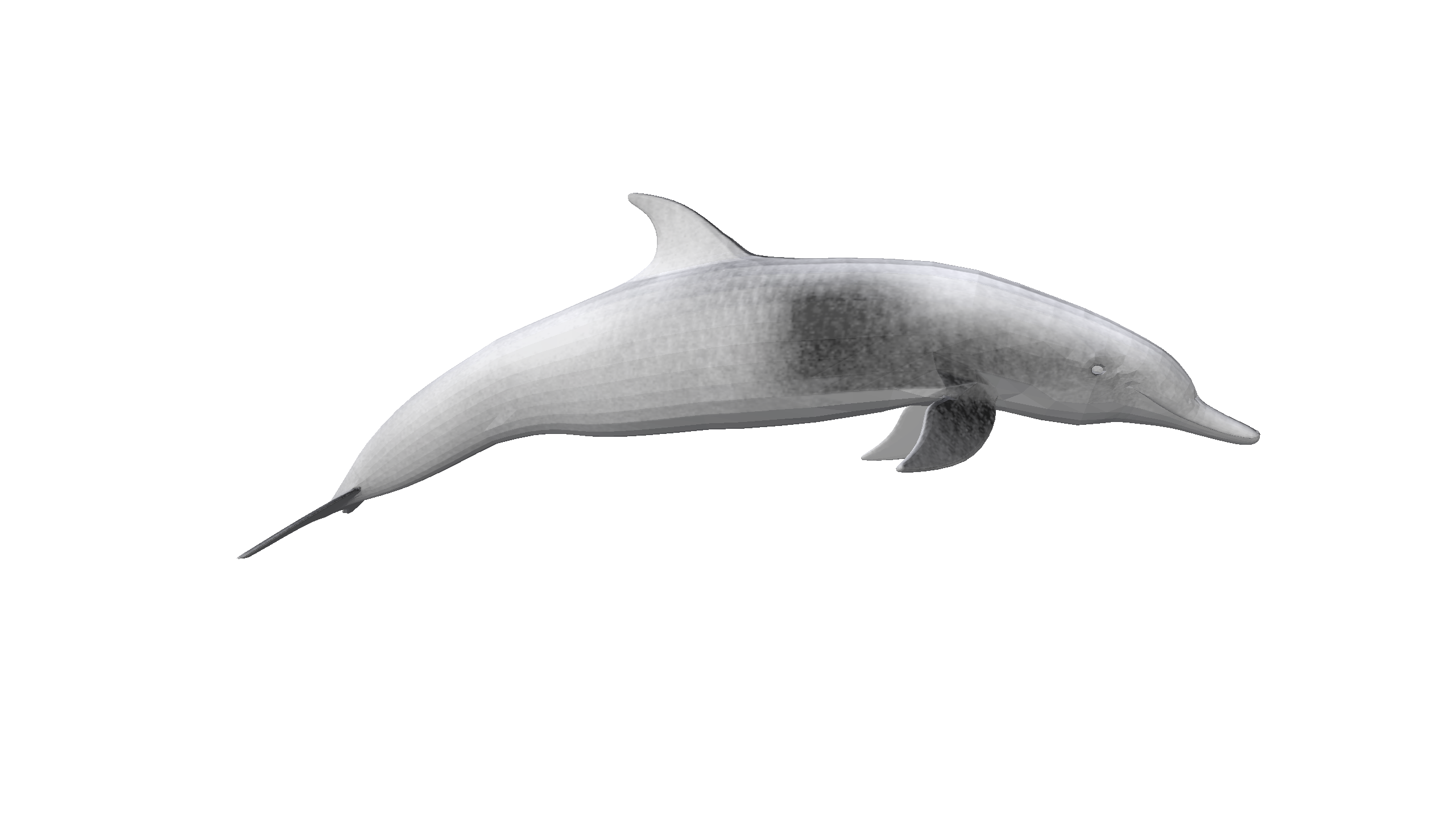}};
        \node[image, right=of hera-mesh] (hera-texture) {\includegraphics[width=.175\linewidth]{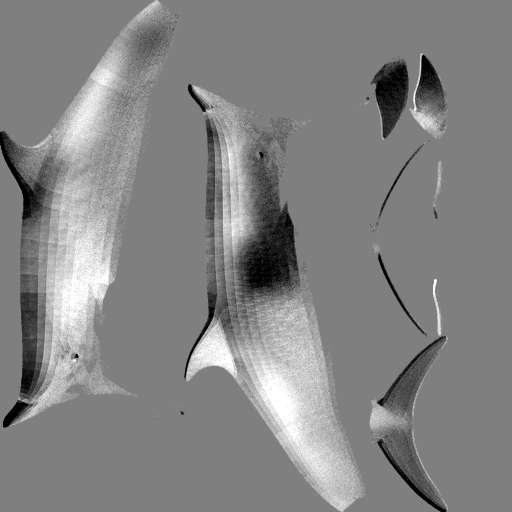}};
        \node[image, right=of hera-texture] (hera-water) {\includegraphics[height=.175\linewidth]{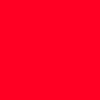}};
        \node[anchor=south west, scale=0.9] at ([xshift=10pt]hera-mesh.south west) {CLL};
        
        \node[image, below=of hera-render] (smurf-render) {\includegraphics[height=.175\linewidth, trim=205 105 195 135, clip]{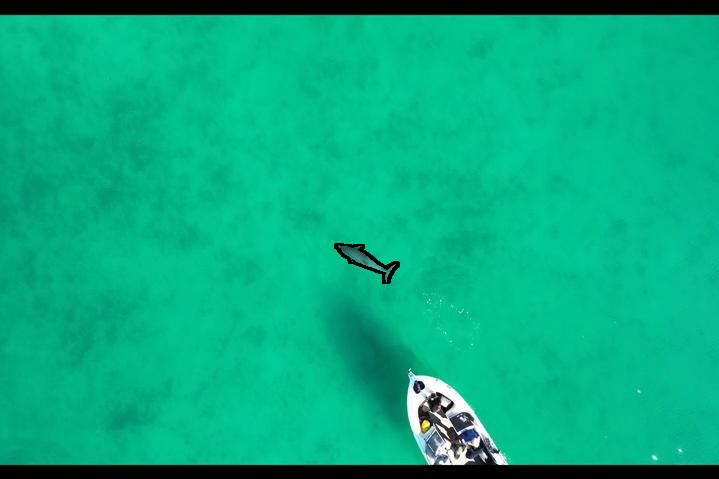}};
        \node[image, right=of smurf-render] (smurf-mesh) {\includegraphics[width=.35\linewidth, trim=50 100 50 90, clip]{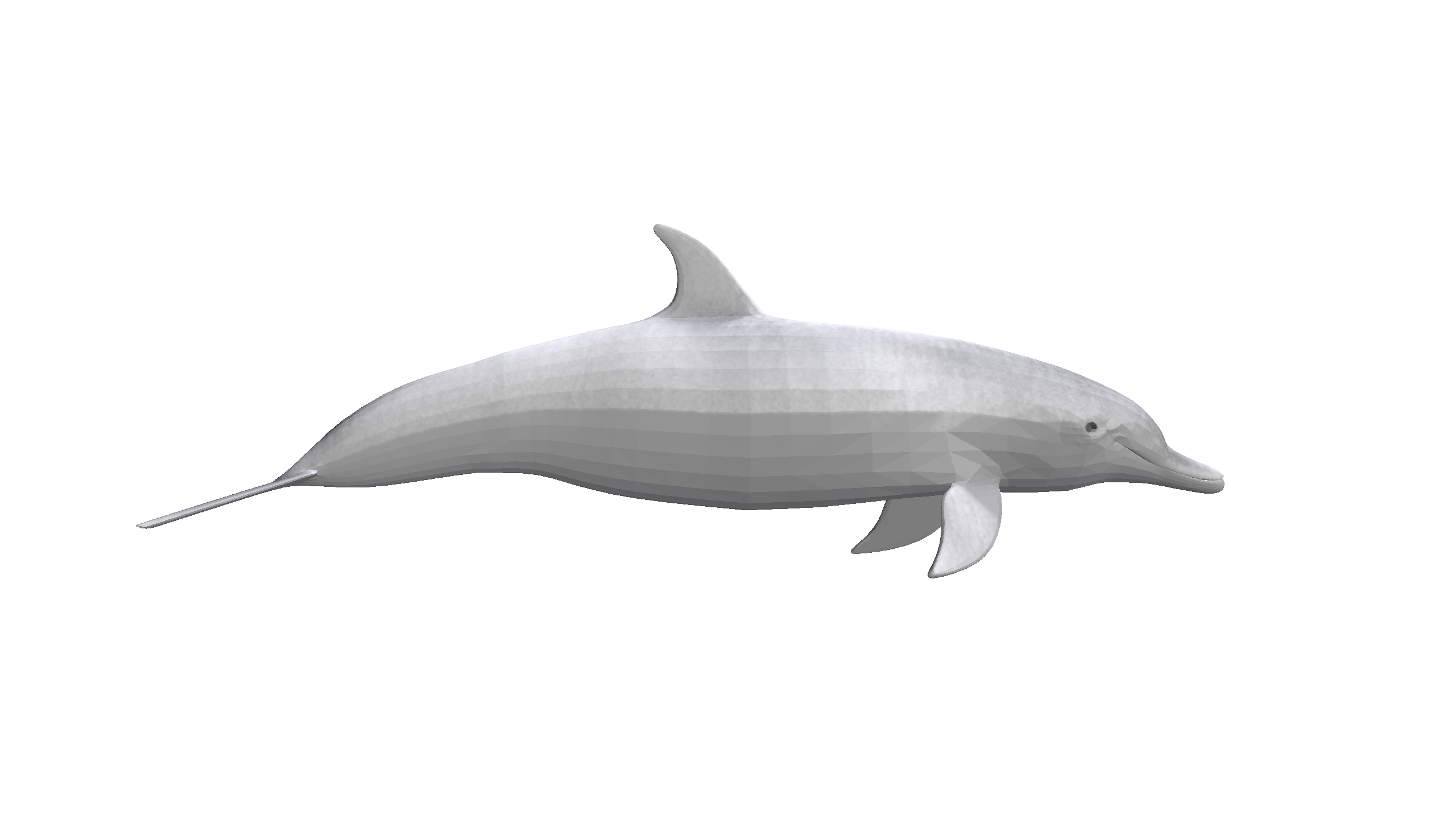}};
        \node[image, right=of smurf-mesh] (smurf-texture) {\includegraphics[width=.175\linewidth]{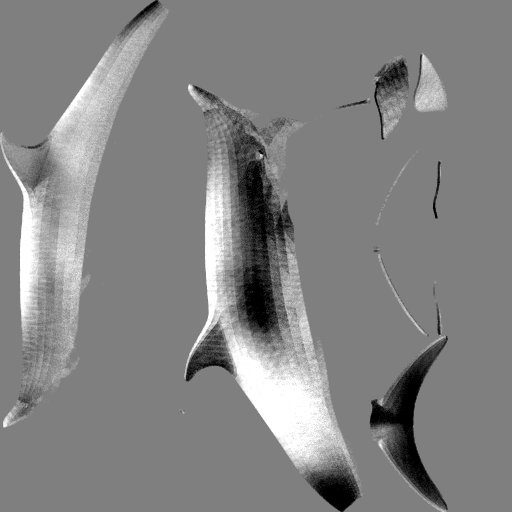}};
        \node[image, right=of smurf-texture] (smurf-water) {\includegraphics[height=.175\linewidth]{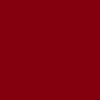}};
        \node[anchor=south west, scale=0.9] at ([xshift=10pt]smurf-mesh.south west) {GRE};
        
        \node[image, below=of smurf-render] (tam-render) {\includegraphics[height=.175\linewidth, trim=215 115 185 125, clip]{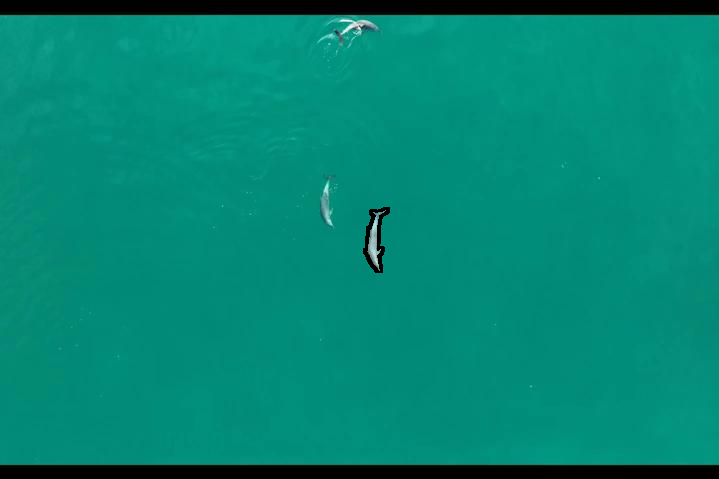}};
        \node[image, right=of tam-render] (tam-mesh) {{\includegraphics[width=.35\linewidth, trim=50 100 50 90, clip]{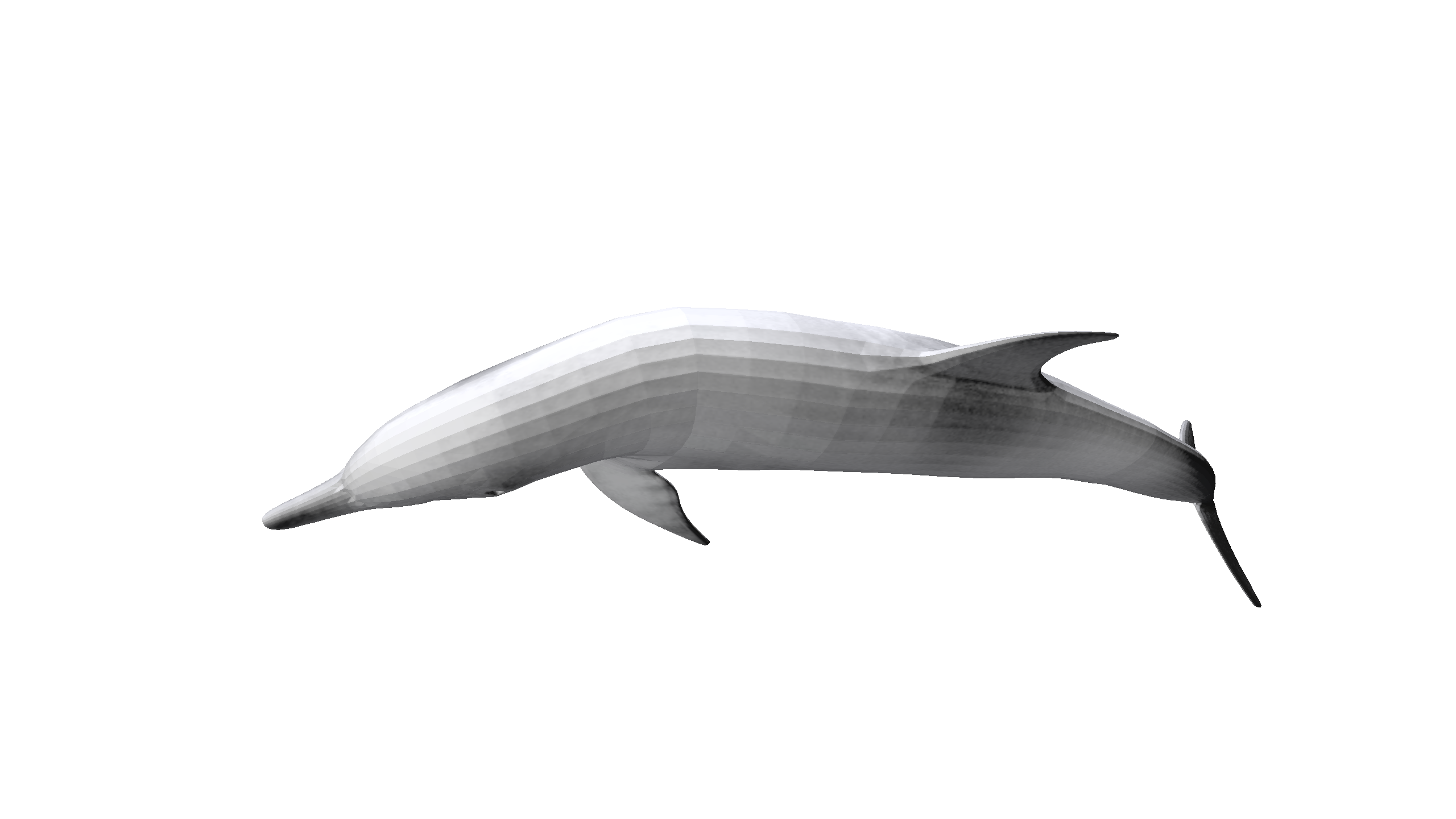}}};
        \node[image, right=of tam-mesh] (tam-texture) {\includegraphics[width=.175\linewidth]{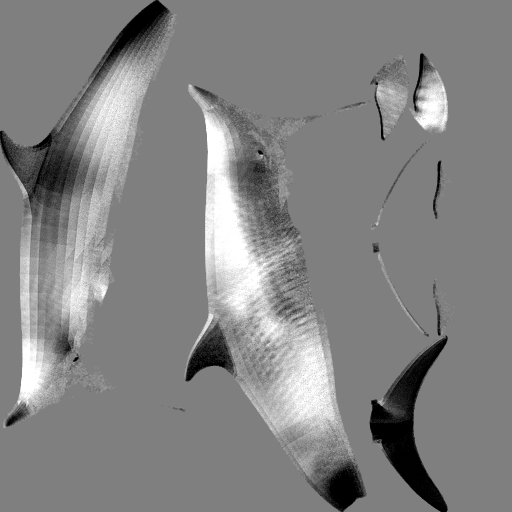}};
        \node[image, right=of tam-texture] (tam-water) {\includegraphics[height=.175\linewidth]{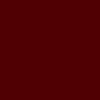}};
        \node[anchor=south west, scale=0.9] at ([xshift=10pt]tam-mesh.south west) {ELM};

        \node[image, below=of tam-render] (guinness-render) {\includegraphics[height=.175\linewidth, trim=245 125 155 115, clip]{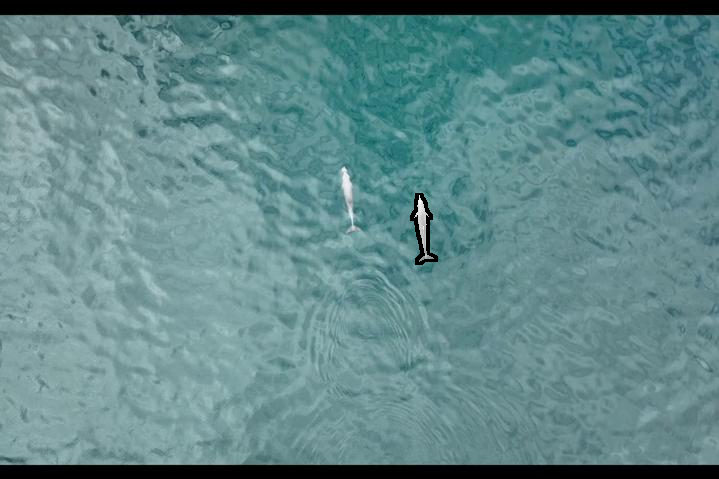}};
        \node[image, right=of guinness-render] (guinness-mesh) {{\includegraphics[width=.35\linewidth, trim=50 100 50 90, clip]{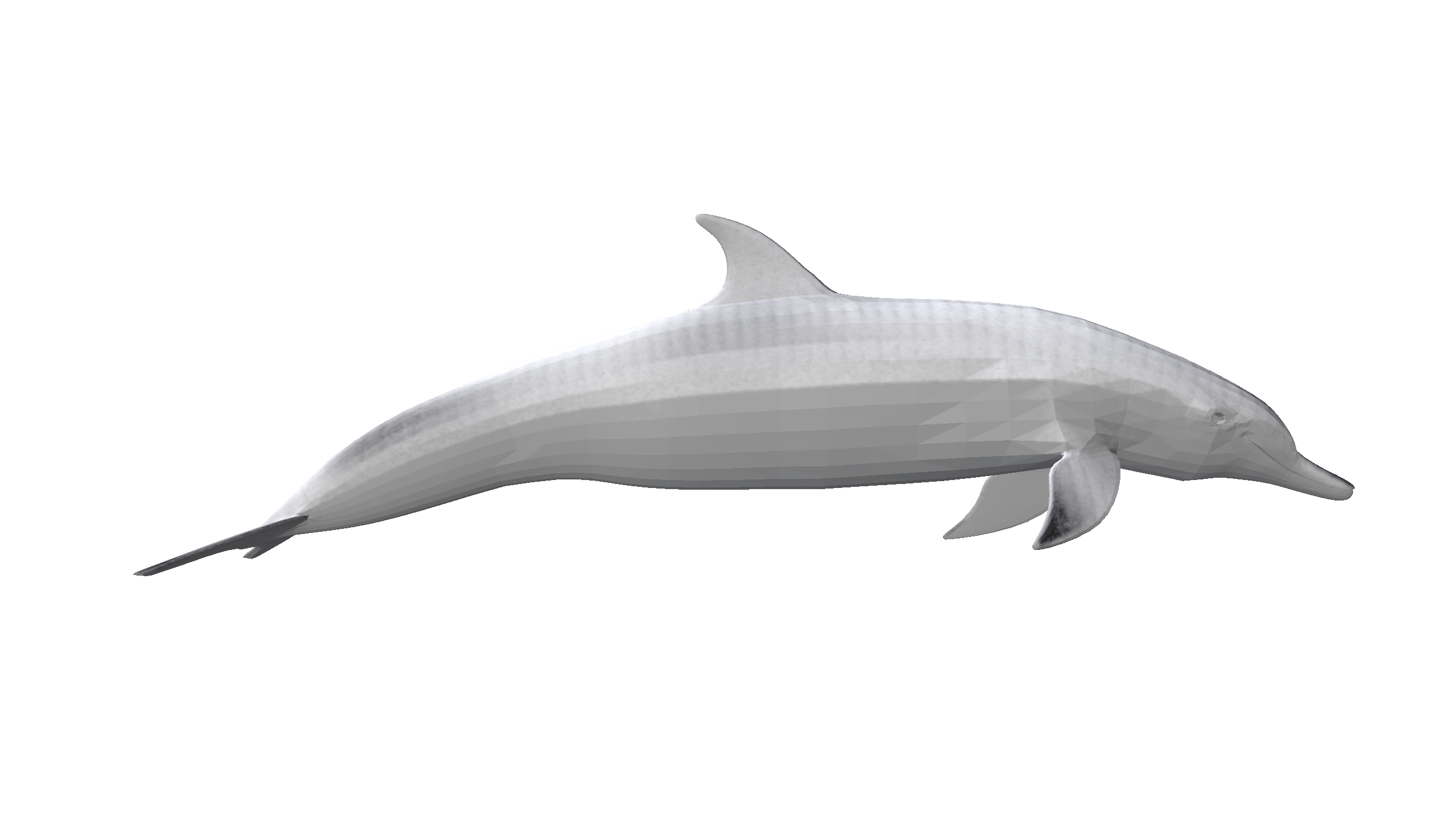}}};
        \node[image, right=of guinness-mesh] (guinness-texture) {\includegraphics[width=.175\linewidth]{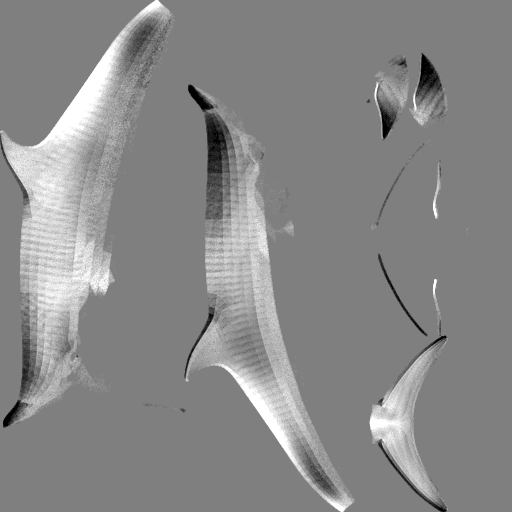}};
        \node[image, right=of guinness-texture] (guinness-water) {\includegraphics[height=.175\linewidth]{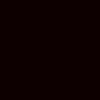}};
        \node[anchor=south west, scale=0.9] at ([xshift=10pt]guinness-mesh.south west) {GUI};

        \node[anchor=center, scale=0.9] (render) at ([yshift=2.5mm]hera-render.north) {$\hat{C}$\strut};
        \node[anchor=center, scale=0.9] (mesh) at ([yshift=2.5mm]hera-mesh.north) {Mesh\strut};
        \node[anchor=center, scale=0.9] (texture) at ([yshift=2.5mm]hera-texture.north) {$I_{\text{albedo}}$};
        \node[anchor=center, scale=0.9] (water) at ([yshift=2.5mm]hera-water.north) {$F_{\text{water}}$};

        \node[anchor=center, scale=0.9] at ($(render)!0.5!(mesh)$) {$=$\strut};
        \node[anchor=center, scale=0.9] at ($(mesh)!0.5!(texture)$) {$+$\strut};
        \node[anchor=center, scale=0.9] at ($(texture)!0.5!(water)$) {$+$\strut};

    \end{tikzpicture}
\caption{\textbf{3D Reconstruction.} Given 4 scenes in our dataset, we show the rendering in a selected frame after optimization, along with the components used for computing it: the dolphin's shape (including localization), its albedo texture, and the water filter.}
\label{fig:3drec}
    
\end{figure}

\paragraph*{Direction Term} When the dolphin moves independently of the drone, we should ensure that it actually moves in its forward direction. We penalize inconsistent behaviour with the cosine distance between the dolphin's forward vector and the normalized change in position. Given $z$ the forward vector of the dolphin in rest pose, we define this loss as
\begin{equation}
    L_{\text{dir}} = \sum_{t=2}^{T} 1 - \left(R_{\bm{\theta}_{t,1}}(z) \cdot \dfrac{\bm{P}_t - \bm{P}_{t-1}}{\lVert \bm{P}_t - \bm{P}_{t-1} \rVert}\right)
\end{equation}
where $R_{\theta}(x)$ applies the 3D rotation with axis angles $\theta$ to the 3D vector $x$. 
When the drone accurately tracks the dolphin (which is often the case), its position should not change, thus we disable this term. Additionally, due to the vertical motion being unobservable from a drone shot, we adjust the vertical position of the dolphin depending on its orientation. For each frame that the dolphin points upwards, we increase its vertical position by $v_{\text{surf}} \;{m}/{s}$, while we decrease it by $v_{\text{dive}} \;{m}/{s}$ when it points downwards.

\begin{figure}[h]
\centering

\begin{tikzpicture} 
    \pgfplotsset{
    every axis/.append style={font=\footnotesize},
    every tick label/.append style={font=\footnotesize}
    }
    \begin{axis}[
    width=.9\linewidth, 
    ylabel={Body volume ($m^3$)},
    xlabel={Body length ($m$)},
    zlabel={Est. mass ($kg$)},
    xmin=1.98, xmax=2.22,
    ymin=0.09, ymax=0.27,
    zmin=95.0, zmax=180.0,
    ytick={0.1,0.15,0.2,0.25},
    xtick={2.0,2.1,2.2},
    ztick={100,118,135,153,170},
    view={60}{30},
    xlabel style={rotate=-36}, 
    ylabel style={rotate=13},
    ]
    
    \pgfplotsset{colormap/violet} 
    \addplot3[surf, opacity=0.6, domain=2.0:2.2, y domain=0.1:0.24]
        {y * (1000.71 - 278.12 * (y - e^(-4.0206 + 2.5929 * ln(x))) / e^(-4.0206 + 2.5929 * ln(x)))};

\end{axis}
\end{tikzpicture}
\caption{\textbf{Mass prediction model.} A 3D visualization of the model we apply to estimate dolphin mass from volume and body length.}
\end{figure}

\subsection{Mass Estimation}
\subsubsection{Body condition}

Following the methods by \citet{christiansen2018maternal}, we calculated body volume ($\text{BV}_{\text{obs}}$) from the dolphins at Sea World Australia. We then estimated the body condition index (BCI) of each animal with the following equation:
\begin{equation}
\text{BCI}_i = \dfrac{\text{BV}_{\text{obs},i} - \text{BV}_{\text{exp},i}}{\text{BV}_{\text{exp},i}}
\end{equation}
where $\text{BV}_{\text{obs},i}$ ($m^3$) represents the measured body volume of dolphin $i$, and $\text{BV}_{\text{exp},i}$ ($m^3$) is the expected body volume based on the log-log linear relationship between body volume and body length ($\text{BL}$):
\begin{equation}
\text{BV}_{\text{exp},i} = \exp\left(-4.02 + 2.59\log\left(\text{BL}_i\right)\right)
\end{equation}

\subsubsection{Body mass and density}

We calculate the dolphins' body mass by multiplying body volume (BV) by body density (BD), which is determined by dividing the body mass measured from the dolphins under human care by their estimated BV. We examined the relationship between BD and body condition index (BCI) using linear regression models. Given that cetaceans store energy primarily as blubber, which is less dense than muscle, we hypothesized a negative BD-BCI relationship: 

\begin{equation}
    \text{BD}_i = \alpha - \beta \times \text{BCI}_i
    \label{eq:bd_bci}
\end{equation}
The predicted body mass (BM) was then calculated as:

\begin{equation}
    \text{BM}_{\text{Pred}} = \text{BV}_{\text{obs}} \times \text{BD}
    \label{eq:bm}
\end{equation}
\begin{equation}
    \text{BM}_{\text{Pred}} = \text{BV}_{\text{obs}} \times \left(\alpha - \beta \times \dfrac{\text{BV}_{\text{obs},i} - \text{BV}_{\text{exp},i}}{\text{BV}_{\text{exp},i}}\right)
\end{equation}

\section{Evaluation}\label{sec:evaluations}

In this section, we present our method's evaluation in comparison with a measurement-based baseline. The values of hyperparameters used in our experiments may be found in the supplementary materials. For all experiments, the model is optimized for 100 epochs, requiring $\sim$20 min per 10 seconds of video on a NVIDIA RTX 4060Ti GPU.

\subsection{Baseline}

To evaluate our method, we implemented a baseline for computing body volume and mass, adapting previous work by \citet{christiansen2019estimating}. The resulting Elliptical Volume Estimation method estimates volume by approximating the animal body with a set of elliptical segments. These segments are obtained by manually annotating selected frames of the videos and then applying a model of width-height ratio learned from animals in captivity (which may result in undesired bias when applied to wild specimen). At each 5\% measurement site along the body, we calculated the height-to-width (HW) ratio for each dolphin under human care \cite{christiansen2019estimating}. Body volume (BV) was then estimated using the segmented elliptical model described by \citet{christiansen2019estimating}:
\begin{multline}
      V_{(s,i)} = BL_{i} \times 0.05 \times \int_{0}^{1} \\
      \frac{\pi \times (W_{A,s,i} + (W_{P,s,i} - W_{A,s,i}) \times x)}{2} \\
  \times \frac{(H_{A,s,i} + (H_{P,s,i} - H_{A,s,i}))}{2} \, dx
  \label{eq:example_label}
\end{multline}
\begin{equation}
    \text{BV}_{\text{obs},i} = \sum_{s=1}^{20} V_{s,i}
    \label{eq:observed_BV}
\end{equation}
Here $V_{s,i}$ represents the volume of segment $s$ (total of 20) for individual $i$, where $BL_{i}$ as body length, and anterior/posterior width ($W_{A,s,i}$, $W_{P,s,i}$) 
 and height ($H_{A,s,i}$, $H_{P,s,i}$) measurements. The integral calculates the elliptical cross-sectional area ($A = \pi \times r_1 \times r_2$) where $r_1$ are the major and minor radii \cite{christiansen2019estimating}.
To account for body tapering towards the tail and rostrum, W and H were set to zero at body endpoints (0\% and 100\% BL). Values at 90\% and 95\% BL were linearly interpolated from 85\% and 100\% BL \cite{christiansen2019estimating}. The total observed body volume ($BV_{(obs,i)}$) for each individual was obtained by summing the volumes of the 20 segments calculated using the first equation.
For wild dolphins, BL and W at the 19 measurement sites were obtained from UAV footage. Since direct H measurements are rarely feasible, H was predicted from the W using the mean HW ratio at each site \cite{christiansen2020variation, christiansen2024prey, christiansen2021poor}.

\begin{table}[t]
\centering
{
\setlength{\tabcolsep}{6pt} 
\renewcommand{\arraystretch}{0.9} 
 
\begin{tikzpicture}
\node[scale=0.9] {
\begin{tabular}{lccccc}
\toprule
\multicolumn{2}{c}{} & \multicolumn{2}{c}{Volume ($m^3$)} & \multicolumn{2}{c}{Mass ($kg$)}\\
\cmidrule(lr){3-4} \cmidrule(lr){5-6}
ID   &  Cond. &  Ellip. &  3D & Ellip. & 3D\\
\midrule
\midrule
        CLL & 0 & 0.122 & 0.106 & 120.55 & 108.81 \\ 
        SLK & 0 & 0.111 & 0.112 & 110.70 & 111.41 \\ 
        ELM & 0 & 0.107 & 0.140 & 111.59 & 136.26 \\ 
        HEA & 0 & 0.144 & 0.162 & 140.19 & 151.93 \\ \midrule
        GRE & 1 & 0.124 & 0.172 & 126.17 & 157.69 \\ 
        ESK & 1 & 0.116 & 0.125 & 114.35 & 120.73 \\ 
        GUI & 1 & 0.104 & 0.135 & 106.98 & 129.27 \\ 
        BUC & 1 & 0.117 & 0.187 & 114.94 & 152.70 \\ \midrule
        SMU & 2 & 0.146 & 0.120 & 142.96 & 123.87 \\ 
        POO & 2 & 0.136 & 0.237 & 132.11 & 167.71 \\ 
        WIM & 2 & 0.121 & 0.117 & 121.41 & 118.24 \\ 
        TAM & 2 & 0.118 & 0.115 & 114.74 & 113.08 \\ \midrule
        WIN & 3 & 0.116 & 0.175 & 128.44 & 158.56 \\ 
        SNR & 3 & 0.120 & 0.120 & 118.49 & 118.31 \\ 
        LAT & 3 & 0.130 & 0.121 & 125.53 & 120.76 \\ 
        SIM & 3 & 0.128 & 0.165 & 127.34 & 150.12 \\ 
\bottomrule
\end{tabular}

};
\end{tikzpicture}
\caption{\textbf{Volume and Mass Estimate}. We report the volume and mass estimate for $16$ subjects across $4$ different Beaufort conditions. We compare the quantities obtained with the Elliptical method with those obtained from the 3D model. Details about the mass estimation and a plot visualization of the results are given in the supplementary material.}
\label{table:evaluation}
}
\end{table}

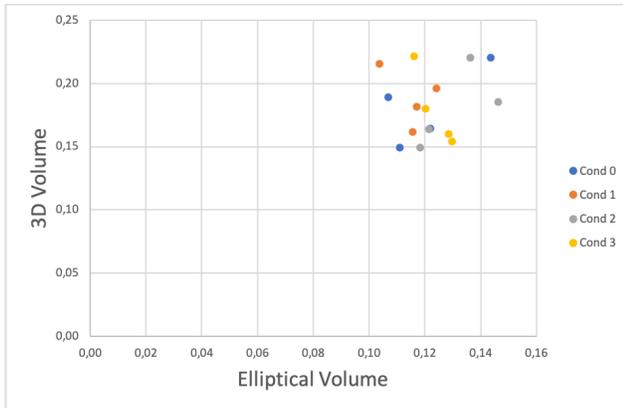
\begin{figure}[t]
    \centering
    \begin{tikzpicture}
    \begin{axis}[
        grid=both,
        legend pos=south west,
        legend style={scale=0.6, nodes={scale=0.6}},
        scaled ticks=false,   
        ticklabel style={/pgf/number format/fixed}, 
        width=\linewidth, 
        height=5cm,
        ylabel={3D Volume},
        xlabel={Elliptical Volume},
        xmin=0.0, xmax=0.25,
        ymin=0.0, ymax=0.25, 
        ytick={0,0.05,0.1,0.15,0.2,0.25},
        xtick={0,0.05,0.1,0.15,0.2,0.25},
        scatter/classes={
            a={mark=*,draw=blue,fill=blue,opacity=0.5},
            b={mark=*,draw=red,fill=red,opacity=0.5},
            c={mark=*,draw=gray,fill=gray,opacity=0.5},
            d={mark=*,draw=orange,fill=orange,opacity=0.5},
            e={mark=*,draw=green,fill=green,opacity=0.5}
        },
    ]
        \addplot [
            scatter,only marks,
            scatter src=explicit symbolic,
        ] coordinates {
            (0.122,0.106)  [a]
            (0.111,0.112)  [a]
            (0.107,0.140)  [a]
            (0.144,0.162)  [a]
            (0.124,0.172)  [b]
            (0.116,0.125)  [b]
            (0.104,0.135)  [b]
            (0.117,0.187)  [b]
            (0.146,0.120)  [c]
            (0.136,0.237)  [c]
            (0.121,0.117)  [c]
            (0.118,0.115)  [c]
            (0.116,0.175)  [d]
            (0.120,0.120)  [d]
            (0.130,0.121)  [d]
            (0.128,0.165)  [d]
            (0.117,0.114)  [e]
            (0.117,0.137)  [e]
            (0.135,0.117)  [e]
            (0.138,0.166)  [e]
            (0.186,0.174)  [e]
            (0.204,0.230)  [e]
            (0.240,0.260)  [e]
        };
    
    \legend{Cond 0,Cond 1,Cond 2,Cond 3,SeaWorld} 
    \end{axis}
    \end{tikzpicture} 
    \caption{\textbf{Results}. We report the results obtained by measuring volume with the 3D body model and the Elliptical Volume method. Different colors indicate different Beaufort conditions.}
    \label{fig:plot}
\end{figure}

\subsection{Discussion}

\paragraph*{In-the-wild Data}
\Cref{fig:rendering} shows renderings of the optimized scene at a selection of frames for multiple scenes in our dataset. These results portray the model's ability of matching the the scene's parameters to the observed mask and color. Nonetheless, the intrinsic under-determination of monocular 3D reconstruction may lead to results such as SIM: while the dolphin shape correctly overlaps with the mask, its pose differs from that showed in the image (where the dolphin is on its side). 
In \Cref{fig:3drec}, on the other hand, we aim to isolate the individual components of our scene model: thus, for each rendered frame, we show the corresponding dolphin geometry, albedo, and water filter. It is interesting to observe that, in the GUI scene, the high transparency in the water results in a near-black filter (very weak filtering with increasing depth). Lastly, we note that most textures present baked shadows, an artifact which could be resolved with a more sophisticated lighting model.
A quantitative evaluation of our method is presented in \Cref{table:evaluation}: we note that in most cases we observe a good correlation with the elliptical method, with our method generally providing larger volume estimates. However, after using them to compute the dolphins' mass, we always achieve reasonable values for the studied species (below 170kg). 
We illustrate the results obtained with the Elliptical Volume estimation method and with the 3D model in Fig.~\ref{fig:plot}.

\begin{table}[t]
\centering
{
\setlength{\tabcolsep}{6pt} 
\renewcommand{\arraystretch}{0.9} 
 
\begin{tikzpicture}
\node[scale=0.95] {
\begin{tabular}{lccccc}
\toprule
& \multicolumn{2}{c}{Volume ($m^3$)} & \multicolumn{3}{c}{Mass ($kg$)}\\
\cmidrule(lr){2-3} \cmidrule(lr){4-6}
ID   &  Ellip. &  3D & Ellip. & 3D & GT\\
\midrule
\midrule
                Huey (1) & 0.117 & 0.114 & 120 & 118 & 124 \\ 
                Huey (2) & 0.117 & 0.137 & 120 & 135 & 124 \\ 
                Tallula  & 0.135 & 0.117 & 137 & 123 & 139 \\ 
                Howie    & 0.138 & 0.166 & 143 & 164 & 154 \\ 
                Evie     & 0.186 & 0.174 & 191 & 182 & 174 \\ 
                Stella   & 0.204 & 0.230 & 200 & 217 & 193 \\ 
                Scooter  & 0.248 & 0.260 & 232 & 240 & 240 \\ 
\bottomrule
\end{tabular}

};
\end{tikzpicture}
\caption{\textbf{SeaWorld Evaluation}. We report the volume and mass estimates for 6 captive subjects. The mass estimates are compared with the known mass values of the examined specimen.}
\label{table:seaworld}
}
\end{table}

\paragraph*{SeaWorld Data}
In order to provide a quantitative evaluation of our method, we performed the same experiments on captive specimen, for which mass measurements are available. The results are reported in~\Cref{table:seaworld}. On average, we obtain an error of 6.96\% relative to the real dolphin's mass. The elliptical baseline, which requires human intervention for manual image annotation, achieves an error of 4.46\% of the real mass on average. From the qualitative results in~\Cref{fig:seaworld}, we observe that our method struggles with smaller specimen, while it provides more realistic estimates on larger ones. This may be imputed to our shape model being independent of any statistics of the real-world shapes of dolphins (such as in SMPL~\cite{loper15smpl1}): while the model can represent potentially any shape, our regularizations are not informed about the distribution of shape and pose parameters. Building a statistical model of shape and pose parameters could be a valuable direction for future research. 

\section{Conclusion}\label{sec:conclusion}
Our work represents the first investigation into metric 3D reconstruction of dolphins from monocular video for assessing body condition. We conducted experiments in-the-wild and with animals under human care, demonstrating that our approach is competitive with a time-consuming manual solution, while being totally automatic. This is a first effort towards accurate and scalable methods for animal monitoring. Future work includes accounting for visual distortions caused by water refraction, collecting additional datasets of captive dolphins with known mass and volume, and constructing statistical models of dolphins shape and pose parameters.  Lastly, while the low resolution in our videos does not seem to affect accuracy, future work could also include extracting bounding boxes from high resolution images, but this would require careful modeling of camera parameters in order to preserve metric properties.

\backmatter

\bmhead{Acknowledgements}
This research was carried out in Guthaaguda, Malgana Sea Country, and we acknowledge the traditional owners of the region. We thank Shark Bay Resources for their continued support and assistance. We also thank Sea World Australia for hosting us and granting access to their animals for data collection. We are also grateful to the dolphin trainers for their dedicated work in training the animals and collecting hand measurements, which made this research possible. 

\section*{Declarations}

\subsection*{Funding}

S.Z. is supported by Fondazione Cariplo e Fondazione CDP within the project MARTA: Model-based ARTtificial intelligence for Animals.

\noindent
This study was supported by grants from the A.H. Schultz Foundation (Dept. of Evolutionary Anthropology, University of Zurich) and the Swiss National Science Foundation (310030\_204974 to M.K.). 

\noindent
This work is partly supported by Sapienza University of Rome via the Seed of ERC grant MINT.AI, cup B83C25001040001.

\begin{appendices}

\section{Hyperparameters}

We provide the values of hyperparameters used in our experiments in \Cref{tab:hyperpar}. Each loss term $L_*^*$ has an associated weight $\lambda_*^*$ used during aggregation. The table is split into the following groups: rendering parameters, loss weights, optimization parameters, and others. The parameters $\frac{1}{\sigma}$, box length and $k$ are not mentioned in the main text, as these are DIB-R parameters which influence how the soft mask is computed. In the optimization group, we split the learning rate (LR) for categories of parameters as we empirically observed benefits in doing so, possibly due to the difference in range of values for color parameters vs transform parameters.

\begin{table}[h]
    \centering
    
    \def\arraystretch{1.3}
    \begin{tikzpicture}
        \node[scale=0.9] {
            \begin{tabular}{c|c}
                \textbf{Parameter} & \textbf{Value} \\ \hline
                $\frac{1}{\sigma}$ & 100000\\
                Box length & 0.01 \\
                $k$ & 40 \\
                $\gamma$ & 17.27 $[mm]$\\
                $\omega$ & 12.29 $[mm]$\\
                $(H_r, W_r)$ & (720, 480) \\
                $(H_a, W_a)$ & (512, 512) \\ \hline
                $L_\text{RGB}$ & 1.0 \\
                $L_\text{mask}$ & 1.0 \\
                $L_\text{pose}$ & 2.0 \\
                $L_\text{scale}$ & 0.001 \\
                $L_\text{scale}^{\text{axes}}$ & 0.1 \\
                $L_\text{scale}^{\text{smooth}}$ & 5.0 \\
                $L_\text{smooth}^{\text{pos}}$ & 500.0 \\
                $L_\text{smooth}^{\text{rot}}$ & 500.0 \\
                $L_\text{smooth}^{\text{pose}}$ & 500.0 \\
                $L_\text{dir}$ (*) & 0.1 \\ \hline
                Epochs & 100 \\
                LR - $\bm{\theta}$ & 0.01 \\
                LR - $\bm{\beta},\bm{P}$ & 0.01 \\
                LR - $I_{\text{albedo}},F_{\text{water}}$ & 0.001 \\
                LR - $S$ & 0.01 \\ \hline
                $v_\text{surf}$ & 6.6 $[m/s]$\\
                $v_\text{dive}$ & 2.2 $[m/s]$  \\
            \end{tabular}
        };
    \end{tikzpicture}
    \caption{Hyperparameter values employed in the experiments described in the main paper. (*): optional parameter, was set to 0 whenever not useful.}
    \label{tab:hyperpar}
\end{table}




\end{appendices}

\newpage
\bibliography{sn-bibliography}

\end{document}